\documentclass[10pt,journal,compsoc]{IEEEtran}
\ifCLASSOPTIONcompsoc
  \usepackage[nocompress]{cite}
\else
  \usepackage{cite}
\fi

\usepackage{amsmath}
\usepackage{amsfonts}
\usepackage{graphicx}
\usepackage{multirow}
\usepackage{epsfig}
\usepackage{setspace}
\usepackage{graphicx}
\usepackage{subfigure}
\usepackage{caption}
\usepackage{color}
\ifCLASSINFOpdf
\else
\fi
\usepackage{url}

\begin{document}
%
\title{Transferable Interactiveness Knowledge for \\Human-Object Interaction Detection}
%
%
%
%

\author{Yong-Lu~Li,
        Xinpeng~Liu,~Xiaoqian~Wu,~Xijie~Huang,~Liang~Xu,~Cewu~Lu,~\IEEEmembership{Member,~IEEE}
\IEEEcompsocitemizethanks{
\IEEEcompsocthanksitem Yong-Lu Li, Xinpeng Liu, Xiaoqian Wu, Xijie Huang, Liang Xu, and Cewu Lu are with the Department of Electrical and Computer Engineering, Shanghai Jiao Tong University, Shanghai, 200240, China. E-mail: \{yonglu\_li, enlighten, otaku\_huang, liangxu\}@sjtu.edu.cn, xinpengliu0907@gmail.com.\protect\\
\IEEEcompsocthanksitem Cewu Lu is the corresponding author, member of Qing Yuan Research Institute and MoE Key Lab of Artificial Intelligence, AI Institute, Shanghai Jiao Tong University, China and Shanghai Qi Zhi institute. E-mail: lucewu@sjtu.edu.cn.
}}

\IEEEtitleabstractindextext{%
\begin{abstract}
Human-Object Interaction (HOI) detection is an important problem to understand how humans interact with objects. 
In this paper, we explore interactiveness knowledge which indicates whether a human and an object interact with each other or not. We found that interactiveness knowledge can be learned across HOI datasets and bridge the gap between diverse HOI category settings. 
Our core idea is to exploit an interactiveness network to learn the general interactiveness knowledge from multiple HOI datasets and perform Non-Interaction Suppression (NIS) before HOI classification in inference.
On account of the generalization ability of interactiveness, interactiveness network is a transferable knowledge learner and can be cooperated with any HOI detection models to achieve desirable results.
We utilize the human instance and body part features together to learn the interactiveness in hierarchical paradigm, i.e., instance-level and body part-level interactivenesses. 
Thereafter, a consistency task is proposed to guide the learning and extract deeper interactive visual clues.
We extensively evaluate the proposed method on HICO-DET,  V-COCO, and a newly constructed PaStaNet-HOI dataset.
With the learned interactiveness, our method outperforms state-of-the-art HOI detection methods, verifying its efficacy and flexibility. Code is available at \url{https://github.com/DirtyHarryLYL/Transferable-Interactiveness-Network}.
\end{abstract}

}

\maketitle

\IEEEdisplaynontitleabstractindextext

%
\IEEEpeerreviewmaketitle

\IEEEraisesectionheading{\section{Introduction}
\label{sec:introduction}}
\IEEEPARstart{H}{uman}-Object Interaction (HOI) detection retrieves human and object locations and infers the interaction classes simultaneously from still image. As a sub-task of visual relationship~\cite{visualgenome,Lu2016Visual}, HOI is strongly related to the human body and object understanding~\cite{fang2018weakly,faster-rcnn,lu2018beyond,maskrcnn}. It is crucial for behavior understanding and can facilitate activity understanding~\cite{activitynet}, imitation learning~\cite{immitation}, etc. Recently, impressive progress has been made by utilizing Deep Neural Networks (DNNs) in this area~\cite{hicodet,Gkioxari2017Detecting,qi2018learning,gao2018ican}.

Generally, humans and objects need to be detected first in HOI detection. Given an image and its detections, humans and objects are often paired exhaustively~\cite{Gkioxari2017Detecting,gao2018ican,qi2018learning}. HOI detection task aims to classify these pairs as various HOI categories. Previous one-stage methods~\cite{hicodet,Gkioxari2017Detecting,gao2018ican,vcoco,qi2018learning} directly classify a pair as specific HOIs. These methods predict \emph{interactiveness} implicitly at the same time, where interactiveness indicates whether a human-object pair is interactive. For example, when a pair is classified as ``eat apple'', we can implicitly predict that it is interactive. 

\begin{figure}[h]
	\begin{center}
		\includegraphics[width=0.48\textwidth]{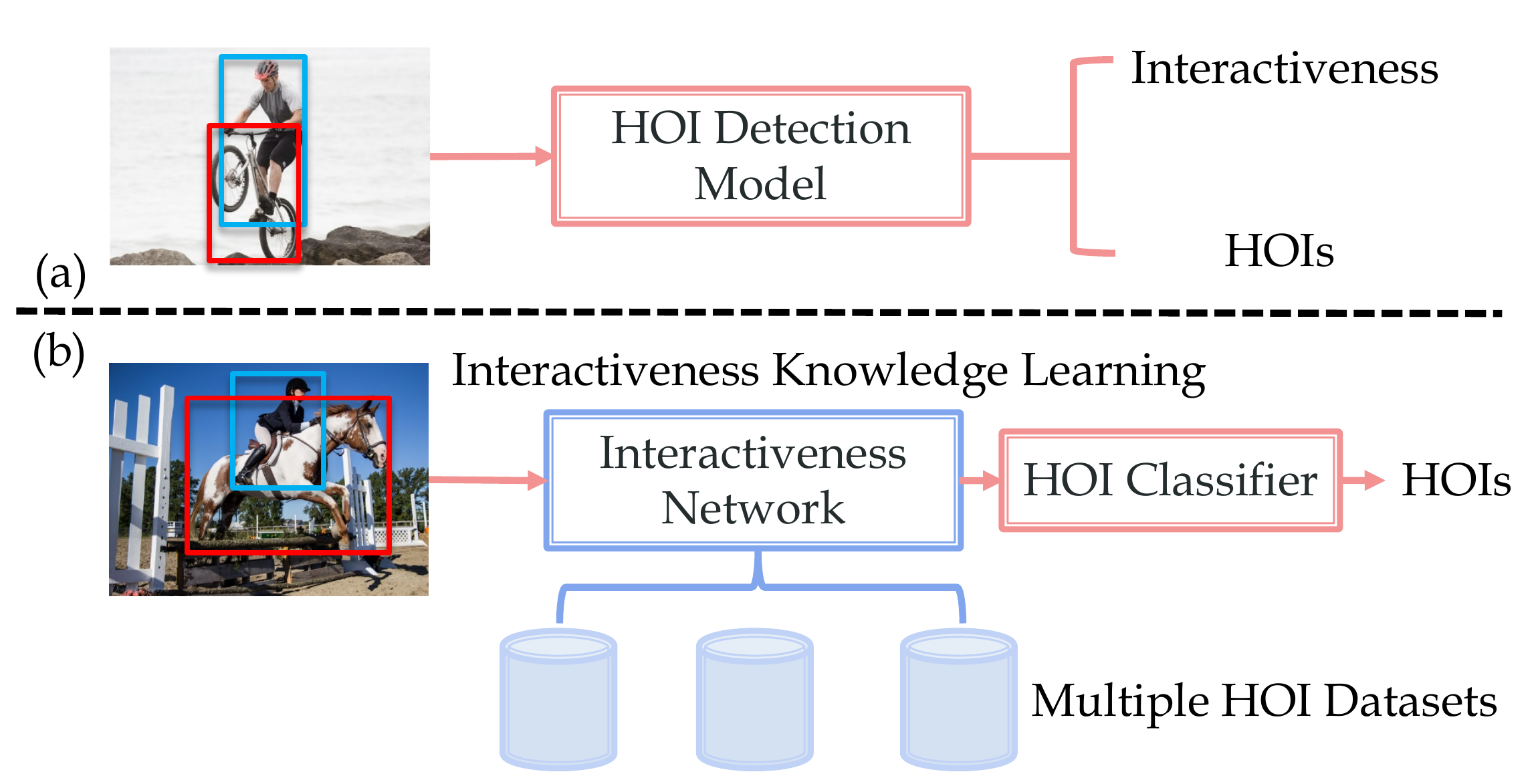}
	\end{center}
	\caption{Interactiveness Knowledge Learning. (a) HOI datasets contain implicit interactiveness knowledge. We can learn it better by performing \textbf{explicit} interactiveness discrimination, and utilize it to improve the HOI detection performance. (b) Interactiveness knowledge is beyond the HOI categories and can be learned across datasets, which can bring greater performance improvement.}
	\label{Figure:pipeline}
	\vspace{-0.3cm}
\end{figure}

Though interactiveness is an essential element for HOI detection, previous methods neglected to study how to utilize it and improve its learning. 
In comparison to various HOI categories, interactiveness conveys more basic information. 
Such an attribute makes it easier to transfer across datasets. 
Based on this inspiration, we propose a \textbf{interactiveness knowledge} learning method as seen in Fig.~\ref{Figure:pipeline}. 
With our method, interactiveness can be learned across datasets and applied to any specific datasets.
By utilizing interactiveness, we take two stages to identify HOIs: first discriminate a human-object pair as interactive or not and then classify it as specific HOIs. 
Compared to the previous one-stage method~\cite{hicodet,Gkioxari2017Detecting,gao2018ican,vcoco,qi2018learning}, we take advantage of powerful interactiveness knowledge that incorporates more information from other datasets. Thus our method can decrease the false positives significantly. Additionally, after the interactiveness filtering in the first stage, we do not need to handle a large number of non-interactive pairs which are overwhelmingly more than interactive ones.

\begin{figure}[!ht]
	\begin{center}
		\includegraphics[width=0.48\textwidth]{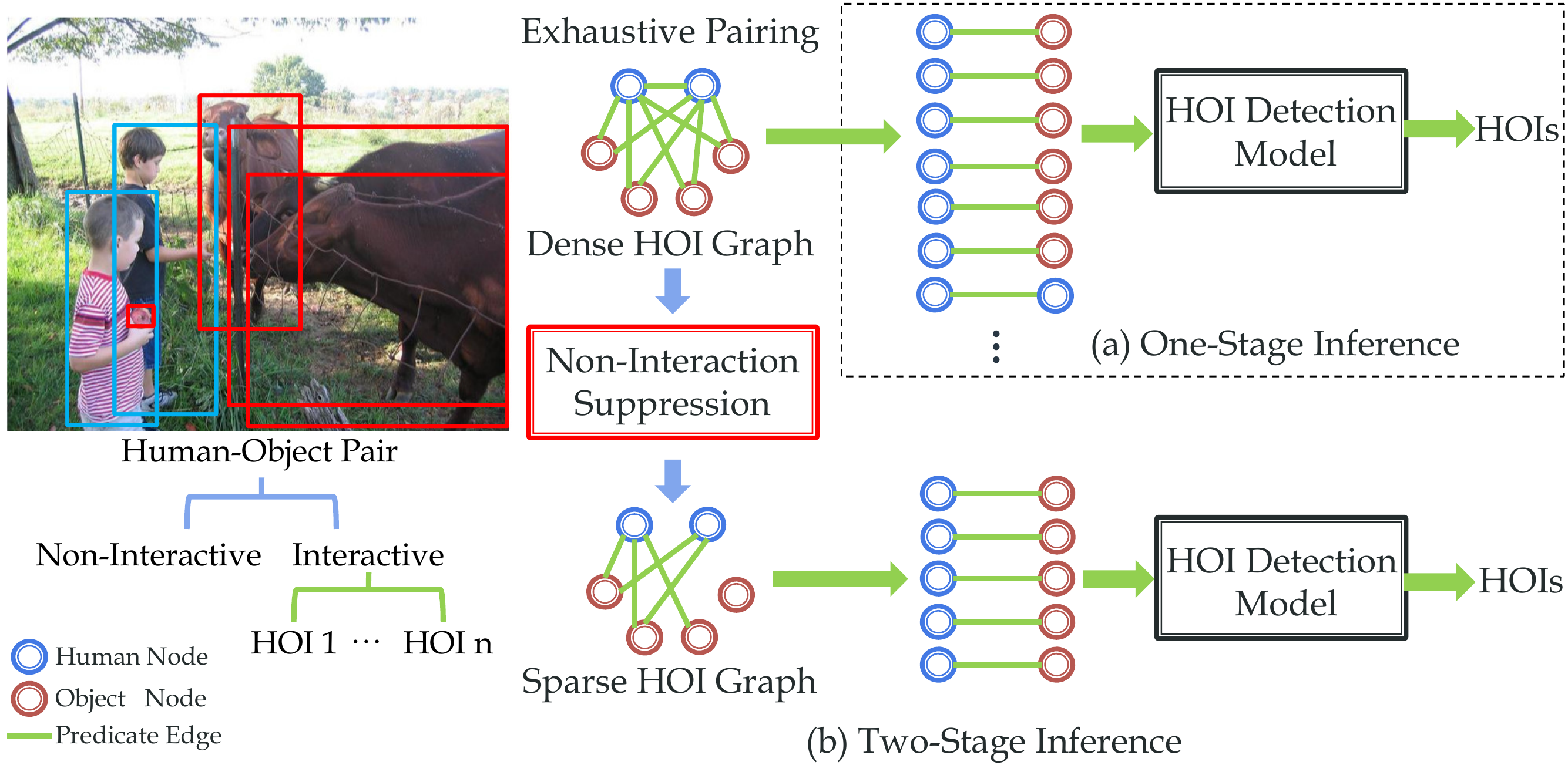}
	\end{center}
	\caption{HOIs within an image can be represented as an HOI graph. Humans and objects can be seen as nodes, whilst the interactions are represented as edges. The exhaustive pairing of all nodes would import overmuch non-interactive edges and do damage to detection performance. Our non-interaction suppression can effectively reduce non-interactive pairs. Thus the dense graph would be converted to a sparse graph and then be classified.}
	\label{Figure:graph-refine}
	\vspace{-0.3cm}
\end{figure}

In this paper, we propose a novel two-stage method to classify pairs hierarchically as shown in Fig.~\ref{Figure:graph-refine}. Our model, Transferable Interactiveness Network (TIN), consists of three networks: Representation Network (extractor, referred to as $\mathbf{R}$), HOI Network (classifier, referred as $\mathbf{C}$), and Interactiveness Network (discriminator, referred as $\mathbf{D}$). 
The interactiveness network $\mathbf{D}$ is creatively utilized for binary classification, i.e., interactive/non-interactive. It benefits the whole model in two aspects.

For one thing, the conventional HOI model is only targeted at HOI detection and classification. Our HOI classifier $\mathbf{C}$ can be trained together with the interactiveness discriminator $\mathbf{D}$ to learn the HOIs and interactiveness knowledge together. 
Under usual circumstances, the ratio of non-interactive edges is dominant within inputs.
Thus, by utilizing binary interactiveness labels converted from HOI labels, the whole model would be trained with a stronger supervised constraint and performs better and more robustly.

For another, noting that the interactiveness network $\mathbf{D}$ only needs binary labels which are beyond the HOI classes, interactiveness is \textbf{transferable} and \textbf{reusable}. Therefore, $\mathbf{D}$ can be used as a transferable knowledge learner to learn interactiveness from multiple datasets and be applied to each of them respectively.

In testing, we adopt the two-stage policy. First, the interactiveness network $\mathbf{D}$ evaluates the interactiveness of a human-object pair (edge) by exploiting the learned interactiveness knowledge, so we can convert the dense HOI graph to a sparse one (Fig.~\ref{Figure:graph-refine}). After this, $\mathbf{C}$ will process the sparse graph and classify the remaining edges.

To implement TIN, we propose a hierarchical framework.
First, we utilize the human/object appearance and spatial configuration as the instance-level features to learn the \textit{interactiveness between instances}.
Second, we further argue that interactiveness has an important characteristic related to human body parts. That said, when interacting with daily objects, only some parts of our body would get involved. For example, in ``read book'', only our head and hands have strong relationships with the book, but not our lower body. We can either stand or lie while reading.
Given this, besides the de facto instance-level features, we further define the interactiveness between object and human body parts, i.e., \textit{part interactiveness}. Then, the human body part feature paired with object feature are used to learn it.
Notably, instance and part interactivenesses have inherent and implicit relationships. 
Their relationship is in line with the Multi-Instance Learning (MIL)~\cite{Maron1998A}, i.e., the instance interactiveness is \textit{false} if and only if all part interactivenesses are \textit{false}. 
To be more explicit, a human is interacting with an object if and only if at least one human part is interacting with the object.  
Thus, when inputting different level features, we can construct this consistency between two levels as an objective in learning.
Moreover, body parts with higher interactiveness scores should be paid more attention to. 
We further use the part attention strategy to strengthen the important parts in HOI inference. 
The experiment (Sec.~\ref{sec:pattern}) verifies our assumption that different HOIs have various part interactiveness patterns. 
For instance, ``ride'' is learned to be more related to feet, thighs, and hands than head and hip. 
Thus, such an attention policy can greatly benefit HOI learning. 

We perform extensive experiments on HICO-DET~\cite{hicodet}, V-COCO~\cite{vcoco}
and a newly constructed dataset PaStaNet-HOI~\cite{hake}. 
Our method cooperated with transferred interactiveness outperforms the state-of-the-art methods by \textbf{1.53} and \textbf{4.35} mAP on the Default set and Rare set of HICO-DET.

\section{Related Works}
\noindent{\bf Visual Relationship Detection.} Visual relationship detection~\cite{Sadeghi2012Recognition,Lu2016Visual,visualgenome,Yatskar2016Situation} aims to detect the objects and classify their relationships simultaneously. In ~\cite{Lu2016Visual}, Lu et al. proposed a relationship dataset VRD and an approach combined with language priors. Predicates within relationship triplet $\langle subject, predicate, object \rangle$ include actions, verbs, spatial and preposition vocabularies. Such vocabulary setting and severe long-tail issue within the dataset make this task quite difficult. Large-scale dataset Visual Genome~\cite{visualgenome} is then proposed to promote the studies in this direction. Recent works ~\cite{xu2017scene,vtranse,yin2018zoom,yang2018graph} put attention on more effective and efficient visual feature extraction and try to exploit semantic information to refine the relationship detection.

\noindent{\bf Human-Object Interaction Detection.} Human-Object Interaction~\cite{Wang2006Unsupervised,Yang2010Recognizing,Ikizler2008Recognizing} is essential to understand human-centric interaction with objects. Several large-scale datasets, such as V-COCO~\cite{vcoco}, HICO-DET~\cite{hicodet}, HCVRD~\cite{hcvrd}, HAKE~\cite{hake} were proposed for the exploration of HOI detection. 
Different from HOI recognition~\cite{Fang2018Pairwise,Delaitre2010Recognizing,hico,Chao2014Predicting,Mallya2016Learning} which is an image-level problem, HOI detection needs to detect interactive human-object pairs and classify their interactions at the instance-level. 
With the assistance of DNNs and large datasets, recent methods have made significant progress~\cite{li2020detailed, kim2020detecting,hou2020visual,zhong2020polysemy,wang2020contextual,kim2020uniondet,gao2020drg,liu2020amplifying,li2020hoi}.

Chao et al.~\cite{hicodet} proposed a multi-stream model combining visual features, spatial locations to help tackle this problem.
To address the long-tail issue, Shen et al.~\cite{Shen2018Scaling} studied zero-shot learning and predicted the verb and object separately.
In InteractNet~\cite{Gkioxari2017Detecting}, an action-specific density map estimation method is introduced to locate interacted objects. 
In~\cite{qi2018learning}, Qi et al. proposed GPNN incorporating DNN and graph model, which uses message parsing to iteratively update states and classifies all possible pairs/edges.
Gao et al.~\cite{gao2018ican} exploited an instance-centric attention module to enhance the information from interest regions and facilitate HOI classification.
Peyre et al. ~\cite{peyre2018detecting} learned a unified space combining visual and semantic language features and detected unseen interactions through entity analogy.

Generally, these methods infer HOI in one-stage and may suffer from severe non-interactive pair domination problems. To address this issue, we utilize interactiveness to explicitly discriminate non-interactive pairs and suppress them before HOI classification.
\begin{figure*}[!ht]
	\begin{center}
		\includegraphics[width=0.9\textwidth]{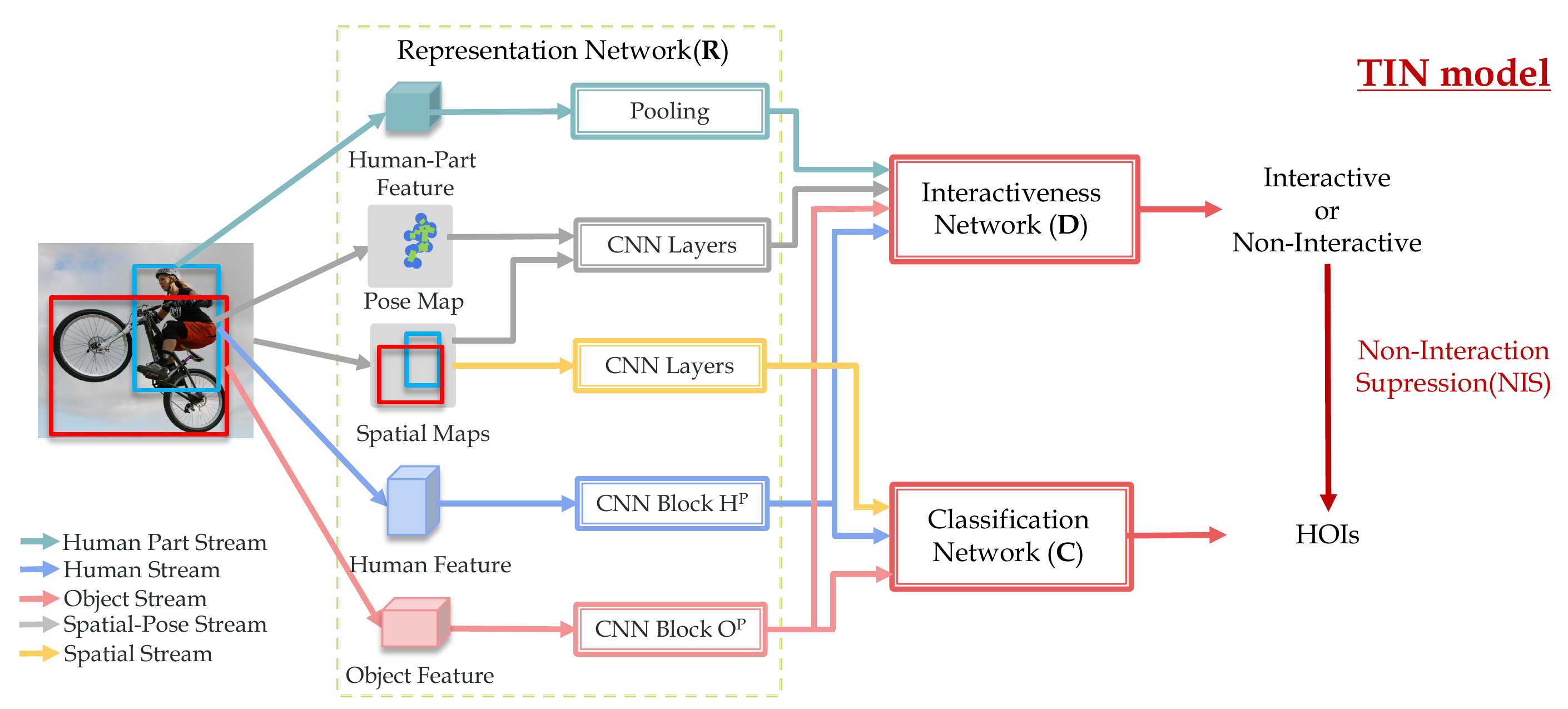}
	\end{center}
	\caption{The overview of our TIN framework. Interactiveness Network $\mathbf{D}$ utilizes interactiveness to reduce false positives caused by overmuch non-interactive pair candidates. 
    Some conventional modules are also included, namely, representation network $\mathbf{R}$ and classification network $\mathbf{C}$. 
    $\mathbf{R}$ is responsible for feature extraction from detected instances. $\mathbf{C}$ utilizes node and edge features to perform HOI classification.
    In testing, $\mathbf{D}$ is utilized in two stages. First, $\mathbf{D}$ evaluates the interactiveness of edges by exploiting the learned interactiveness knowledge and impose NIS on $\mathbf{C}$. Second, combined with interactiveness score from $\mathbf{D}$, $\mathbf{C}$ will process the sparse graph and classify the remaining edges.}
	\label{Figure:overview}
\end{figure*}

\noindent{\bf Part-based Action Recognition.} Part-level human feature takes further insight into human-object interaction. 
Based on the whole person and part bounding boxes, Gkioxari et al.~\cite{Gkioxari2014Actions} developed a part-based model to make fine-grained action recognition. 
In~\cite{Fang2018Pairwise}, Fang et al. proposed a pairwise body part attention model that can learn to focus on crucial parts and their correlations. An attention-based feature selection method and a pairwise parts representation learning scheme are introduced. 
In our work, we utilize the body part features and whole body feature to learn hierarchical interactivenesses. And the unique consistency between the two levels is fully explored to guide the learning.

\section{Preliminary}
HOI representation can be described as a graph model~\cite{qi2018learning, xu2017scene} as seen in Fig.~\ref{Figure:graph-refine}. Instances and relations are expressed as nodes and edges respectively. With exhaustive pairing~\cite{Gkioxari2017Detecting,gao2018ican}, HOI graph $\mathcal{G} = (\mathcal{V}, \mathcal{E})$ is dense connected, where $\mathcal{V}$ includes human node $\mathcal{V}_h$ and object node $\mathcal{V}_o$. 
Let $v_h \in \mathcal{V}_h$ and $v_o \in \mathcal{V}_o$ denote the human and object nodes. Thus edges $e \in \mathcal{E}$ are expressed as $e = (v_h, v_o) \in \mathcal{V}_h \times \mathcal{V}_o$. With $n$ nodes, exhaustive paring will generate a mass of edges. We aim to assign HOI (including no HOI) labels on those edges. Considering that a vast majority of non-interactive edges existing in $\mathcal{E}$ should be discarded, our goal is to seek a sparse $\mathcal{G}^{*}$ with corrected HOI labeling on its edges.

\section{Our Method}
\subsection{Overview}
In this section, we introduce \textbf{interactiveness knowledge} to advance HOI detection performance. That is, explicitly discriminate the non-interactive pairs and suppress them before HOI classification. From the semantic point of view, interactiveness provides more general information than conventional HOI categories. Since any human-object pair can be assigned binary interactiveness labels according to the HOI annotations, i.e., ``interactive'' or ``non-interactive'', interactiveness knowledge can be learned from multiple datasets with different HOI category settings and transferred to any specific datasets.

The overview of our TIN framework is shown in Fig.~\ref{Figure:overview}. We propose interactiveness network $\mathbf{D}$ (interactiveness discriminator) which utilizes interactiveness to reduce false positives caused by overmuch non-interactive pair candidates. 
Conventional modules are also included, namely, representation network $\mathbf{R}$ (feature extractor) and classification network $\mathbf{C}$ (HOI classifier). 
$\mathbf{R}$ is responsible for feature extraction from detected instances. $\mathbf{C}$ utilizes node and edge features to classify HOIs.
In testing, $\mathbf{D}$ is utilized in two stages. First, $\mathbf{D}$ evaluates the interactiveness of edges by exploiting the learned interactiveness knowledge, so we can convert the dense HOI graph to a sparse one. Second, combined with interactiveness score from $\mathbf{D}$, $\mathbf{C}$ will process the sparse graph and classify the remaining edges. 

In subsequent sections, we first introduce the conventional modules $\mathbf{R}$ and $\mathbf{C}$ in Sec.~\ref{sec:rc}. Then, the structure of the interactiveness network $\mathbf{D}$ is detailed in Sec.~\ref{sec:p}. Finally, the Non-Interaction Suppression (NIS) is discussed in Sec.~\ref{sec:test}.

\begin{figure*}[!ht]
	\begin{center}
		\includegraphics[width=0.9\textwidth]{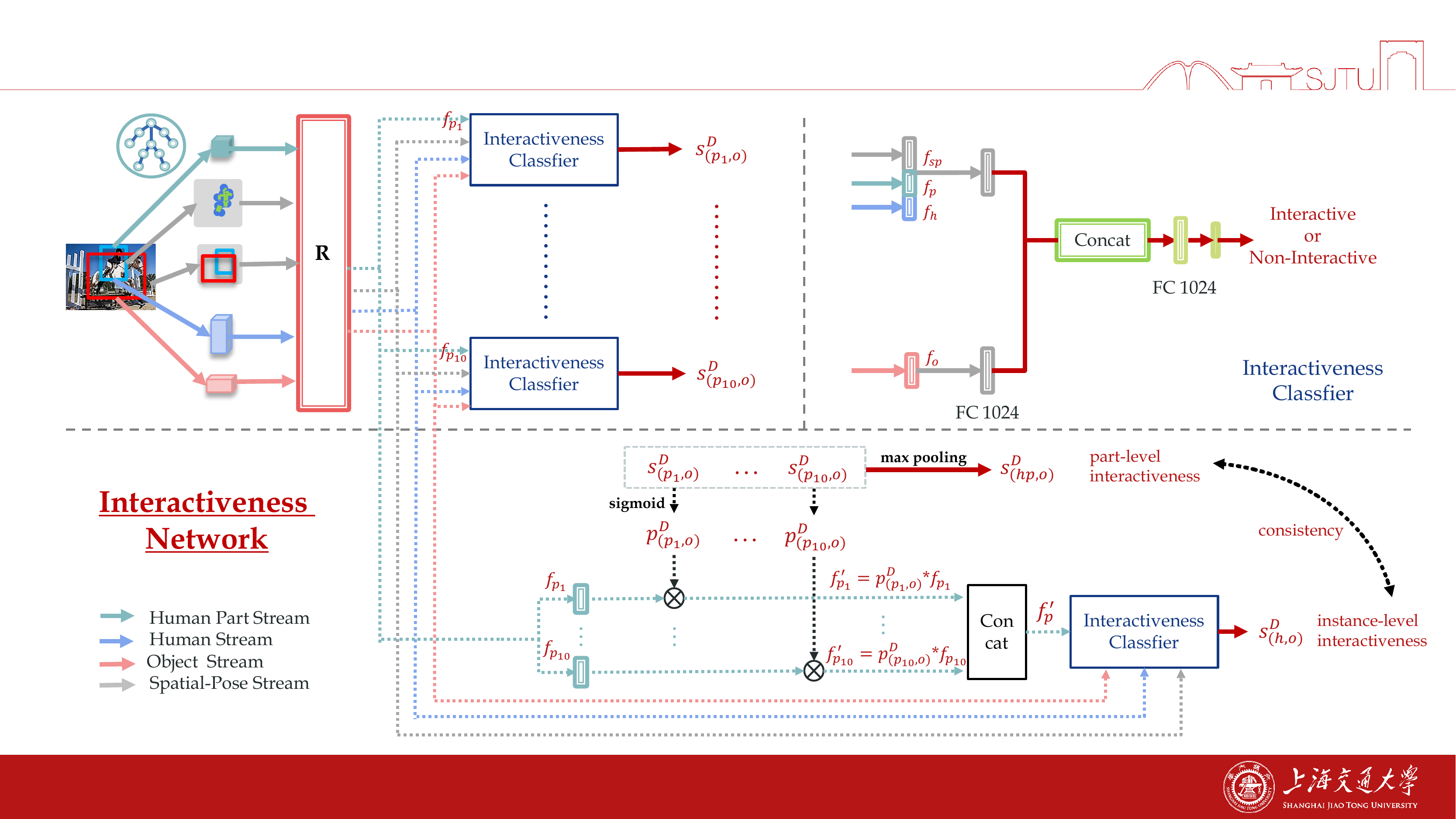}
	\end{center}
	\caption{The architeture of interactiveness network $\mathbf{D}$. 
    There are eleven interactiveness binary classifiers in $\mathbf{D}$, i.e., ten for part interactivenesses and one for instance interactiveness (illustrated in the upper right part). For part-level classifier, the $i$-th part feature $f_{p_i}$ together with $f_h$, $f_o$ and $f_{sp}$, are concatenated and input to FCs and Sigmoid to generate part interactiveness probability $p_{(p_i,o)}^\mathbf{D}$. 
    Meanwhile, we utilize part interactivenesses to select important parts, i.e.  $f_{p_i}^{'}=p_{(p_i,o)}^\mathbf{D} * f_{p_i}$. 
    For instance-level classifier, we concatenate ten re-weighted part features $f_{p_i}^{'}$ ($1 \leq i \leq 10$) and input them to FCs to generate instance interactiveness score $s^{\mathbf{D}}_{(h,o)}$.
    Finally, consistency between two levels is constructed as the objective.}
	\label{Figure:overview_D}
	\vspace{-0.3cm}
\end{figure*}

\subsection{Representation and Classification Networks}
\label{sec:rc}
First, we make a brief introduction to the representation network $\mathbf{R}$ and classification network $\mathbf{C}$. 

\noindent{\bf Human and Object Detection.} In HOI detection, humans and objects need to be detected first. In this work, we follow the setting of~\cite{gao2018ican} and employ the Detectron~\cite{detectron} with ResNet-50-FPN~\cite{fpn} to prepare bounding boxes and detection scores. Before post-processing, detection results will be filtered by the detection score thresholds first.

\noindent{\bf Representation Network.} In previous methods~\cite{hicodet,Gkioxari2017Detecting,gao2018ican}, $\mathbf{R}$ is often modified from object detector such as Faster R-CNN~\cite{faster-rcnn}. We also exploited a Faster R-CNN~\cite{faster-rcnn} with ResNet-50~\cite{resnet} based $\mathbf{R}$ here. During training and testing, $\mathbf{R}$ is frozen and acts as a feature extractor.  
Given the detected boxes, we produce human and object features by cropping ROI pooling feature maps according to box coordinates. 

\begin{figure}[!ht]
	\begin{center}
		\includegraphics[width=0.48\textwidth]{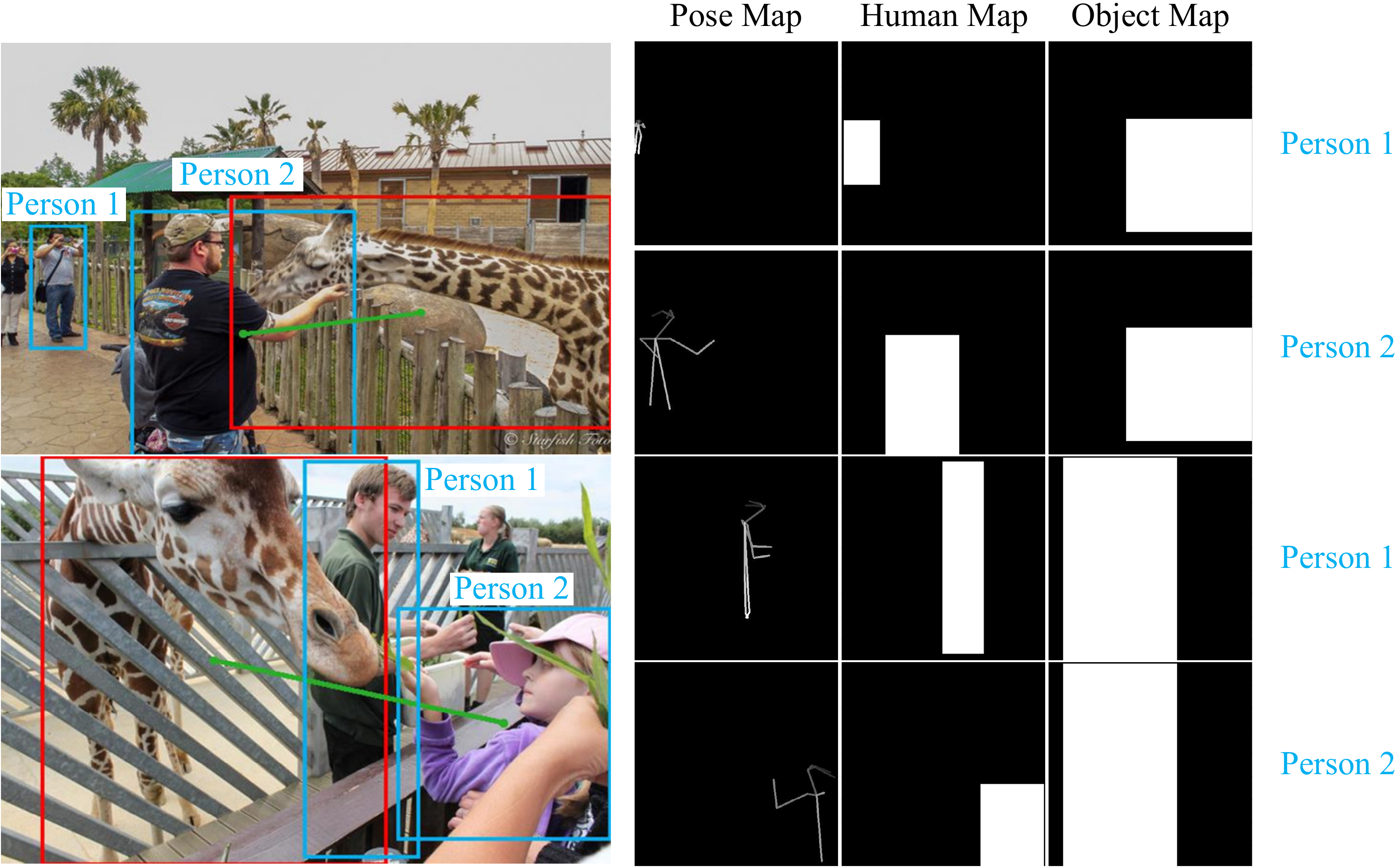}
	\end{center}
	\caption{Inputs of the spatial-pose stream. Three kinds of maps are included: pose map, human map and object map. Person 2 in two images both have interaction ``feed'' with giraffes. But two pairs of Person 1 and giraffe are all non-interactive. Their poses and locations are helpful for the interactiveness discrimination.}
	\label{Figure:d work reason}
	\vspace{-0.3cm}
\end{figure}

\noindent{\bf Classification Network.} As for $\mathbf{C}$, multi-stream architecture and late fusion strategy are frequently used and proved effective~\cite{hicodet,gao2018ican}. 
Following~\cite{hicodet,gao2018ican}, for our classification network $\mathbf{C}$, we utilize a human stream and an object stream to extract human, object, and context features. Within each stream, a residual block~\cite{resnet}  (denoted as $H^{\mathbf{C}}$ and $O^{\mathbf{C}}$) with pooling layer and fully connected layers (FCs) are adopted.
Moreover, an extra spatial stream~\cite{hicodet} is adapted to encode the spatial locations of instances. Its input is a two-channel tensor consisting of a human map and an object map, shown in Fig.~\ref{Figure:d work reason}. Human and object maps are all 64x64 and obtained from the human-object union box. In the human channel, the value is 1 in the human box and 0 in other areas. The object channel is similar which has a value of 1 in the object box and 0 elsewhere.
Following the late fusion strategy, each stream will first perform HOI classification. The prediction scores of human and object streams will be fused by the element-wise sum in the same proportion, then we multiply it with the score of the spatial stream and produce the final result of $\mathbf{C}$.

\subsection{Interactiveness Network}
\label{sec:p}
The interactiveness network $\mathbf{D}$ is designed for binary classification: interactive/non-interactive.
As aforementioned, we hierarchically infer the interactiveness: 
1) \textbf{Instance-Level.} The human/object appearance and spatial configuration are used as the instance-level features to predict the interactiveness between human and object. 
2) \textbf{Part-Level.} Notably, we further utilize the human body part features to take a deeper insight into the interactivenesses between body parts and object.
With interactivenesses from two levels, we then can use the consistency between them to guide the learning.
To summarize, there are four kinds of streams (human, object, spatial-pose, and part) in $\mathbf{D}$. 
Each of them focuses on different elements of the HOIs in images. The architecture of interactiveness network $\mathbf{D}$ is shown in Fig.~\ref{Figure:overview_D}.

In subsequent sections, we illustrate the details of the interactiveness network. 
First, we introduce three streams based on instance-level features in Sec.~\ref{sec:d_stream}. Second, we explore the part stream based on part-level features in Sec.~\ref{sec:d_part}. 
Third, we detail the interactiveness binary classification via four streams in Sec.~\ref{sec:binary-classifier}. 
Next, consistency between two interactive levels is discussed in Sec.~\ref{sec:d_ca}. 
Finally, a low-grade instance suppressive function is further proposed in Sec.~\ref{sec:lis}. 

\subsubsection{Three Streams with Instance-Level Features}
\label{sec:d_stream}
Interactiveness needs to be learned by extracting and combining essential information. The visual appearance of humans and objects are obviously required.
Besides, interactive and non-interactive pairs also have other distinguishing features, e.g. spatial location and human pose information.
For example, in the upper image of Fig.~\ref{Figure:d work reason}, Person 1 and the giraffe far away from him are not interactive. Their spatial maps~\cite{hicodet} can provide pieces of evidence to help with classification. 
Furthermore, pose information is also helpful. In the lower image, although two people are both close to the giraffe, only Person 2 and the giraffe are interactive. The arm of Person 2 is uplift and touching the giraffe. Whilst Person 1 is back on to the giraffe, and his pose is quite different from the typical pose of ``feed''.

Given this, we argue that the combination of visual appearance, spatial location and pose information is the key to interactiveness discrimination.
Hence, $\mathbf{D}$ needs to encode all these key elements together to learn interactiveness knowledge. A natural choice is a multi-stream architecture as presented: human, object, and spatial-pose streams, and the instance-centric attention module~\cite{gao2018ican} is also adopted. 

\noindent{\bf Human and Object Streams.}
\label{sec:d_ho}
For human and object appearance, we extract ROI pooling features from $\mathbf{R}$, then input them into residual blocks $H^{\mathbf{D}}$ and $O^{\mathbf{D}}$. The architecture of $H^{\mathbf{D}}$ and $O^{\mathbf{D}}$ is the same as $H^{\mathbf{C}}$ and $O^{\mathbf{C}}$ (Sec.~\ref{sec:rc}). Through subsequent global average pooling and FCs, the output features of two streams are denoted as $f_h$ and $f_o$, respectively.

\noindent{\bf Spatial-Pose Stream}
Different from~\cite{hicodet}, our spatial-pose stream input includes a special 64x64 pose map.
Given the union box of each human and the paired object, we employ pose estimation~\cite{fang2017rmpe} to estimate 17 body keypoints (in COCO format~\cite{coco}). Then, we link the keypoints with lines of different gray values ranging from 0.15 to 0.95 to represent different body parts, which implicitly encodes the pose features. Whilst the other area is set as 0.
Finally, we reshape the union box to 64x64 to construct the pose map. 

We concatenate the pose map with human and object maps which are the same as those in the spatial stream of $\mathbf{C}$. This forms the input for our spatial-pose stream.
Next, we exploit two convolutional layers with max-pooling and two 1024 sized FCs to extract the feature $f_{sp}$ of three maps. Last, the output will be concatenated with the outputs of other streams for the next interactiveness discrimination.

\subsubsection{Part Stream with Part-Level Features}
\label{sec:d_part}
Conventional methods usually focus on instance interaction recognition, while body part-object interaction is often overlooked. 
This is partially caused by the difficulty to annotate the massive relationships between body parts and objects. However, such a relationship can be more easily defined via interactiveness. Therefore, we explore to adopt the human body part features in interactiveness learning.

\begin{figure}[!ht]
	\begin{center}
		\includegraphics[width=0.45\textwidth]{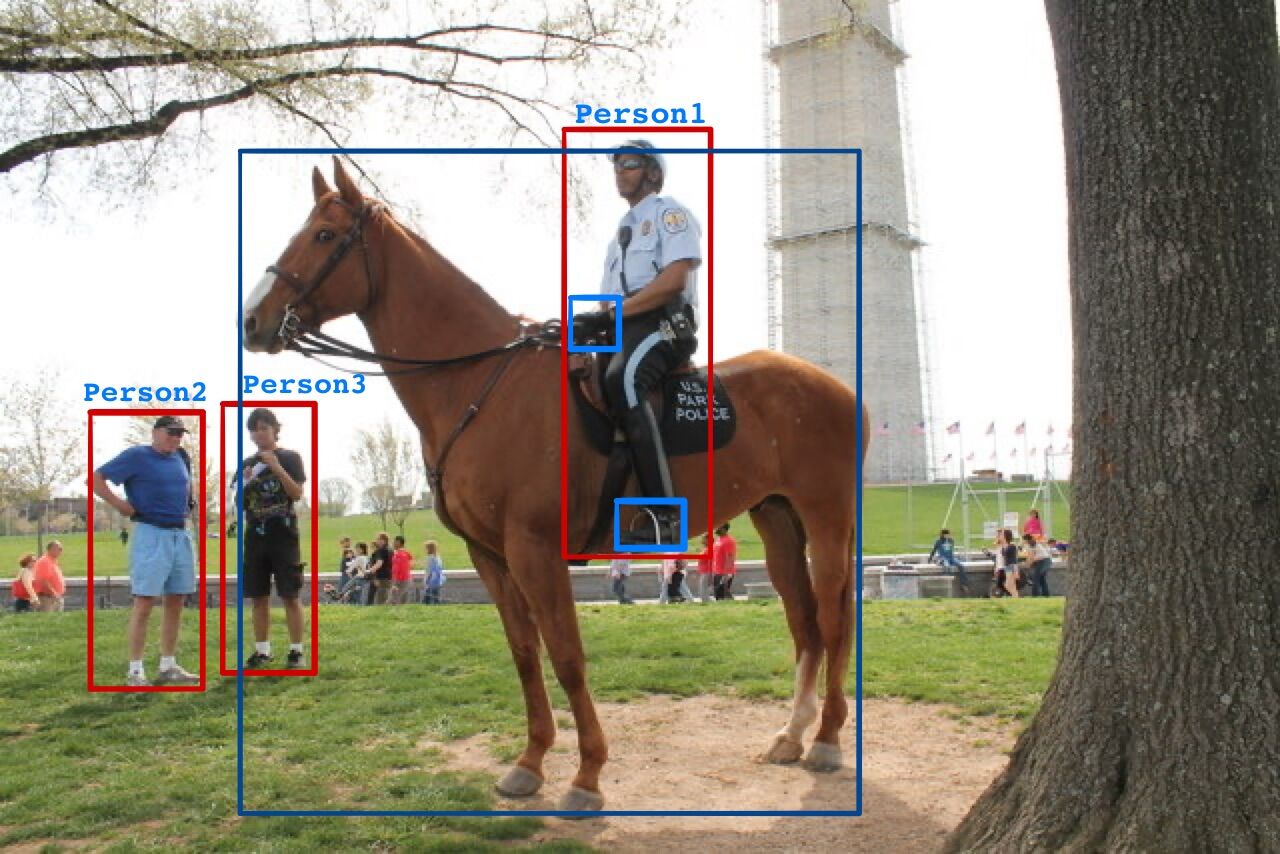}
	\end{center}
	\caption{The illustration of the relationship between part-level and instance-level interactiveness. Person\ 1 is inferred to interact with the horse from the features of his hands and feet, while Person\ 2 and Person\ 3 are not interacting with the horse based on the same reason.}
	\label{Fig:local_ex}
	\vspace{-0.3cm}
\end{figure}

To locate the parts, we first use the pose estimation~\cite{fang2017rmpe} to construct ten part boxes (Fig.~\ref{Fig:local_ex}) following~\cite{Fang2018Pairwise}, i.e., head, upper arms, hands, hip, thighs, and feet. Each part box is centered with a corresponding detected joint. And the size of a part box is decided by scaling the distance between the neck and pelvis joints. 
Second, for part stream, we extract the ROI pooling feature from the detected part box as the part feature, i.e., $f_{p_i}$ for the $i$-th part, $1 \leq i \leq 10$.

\subsubsection{Binary Interactiveness Classifier}
\label{sec:binary-classifier}
There are eleven compact interactiveness binary classifiers (``Interactiveness Classifier'' in Fig.~\ref{Figure:overview_D}) with similar structure in $\mathbf{D}$, i.e., ten for part interactivenesses and one for instance interactiveness. 
They all take four kinds of features from the four aforementioned streams as inputs and are constructed by simple concatenate operation and FC layers.
The detailed structure of the interactiveness classifier is illustrated in the upper right part of Fig.~\ref{Figure:overview_D}.

For the \textbf{part-level classifier}, the $i$-th part feature $f_{p_i}$ together with $f_h$, $f_o$ and $f_{sp}$, are concatenated and input to FCs and Sigmoids to generate the part interactiveness probability $p_{(p_i,o)}^\mathbf{D}=Sigmoid(s^{\mathbf{D}}_{(p_i,o)})$, where $s^{\mathbf{D}}_{(p_i,o)}$ indicates the part interactiveness score of the $i-th$ part.
Particularly, fine-grained part interactiveness also has another characteristic, i.e. \textbf{sparsity}. 
For example, in Fig.~\ref{Fig:local_ex}, to judge whether a person is riding a horse, certain parts should be paid more attention. That is, hands, hip, feet seem more important than head and upper arms.
Thus, utilizing part interactiveness as attention is a natural choice to select the important parts, i.e.  $f_{p_i}^{'}=p_{(p_i,o)}^\mathbf{D} * f_{p_i}$. 
After re-weighting, the information transported to the next instance-level classifier will be filtered. Thus, the model can focus on more important parts and ignore the possible noise imported by the other parts.

For the \textbf{instance-level classifier}, we concatenate ten re-weighted part features $f_{p_i}^{'}$ ($1 \leq i \leq 10$) as $f_{p}^{'}$. Then, similar to the part-level, we input $f_{sp}, f_{p}^{'}, f_h, f_o$ to a classifier and generate the instance interactiveness score $s^{\mathbf{D}}_{(h,o)}$.
At last, we obtain the instance interactiveness probability $p_{(h,o)}^\mathbf{D}=Sigmoid(s^{\mathbf{D}}_{(h,o)})$. 
With $p_{(h,o)}^\mathbf{D}$ and the binary labels converted from HOI labels, we can construct the binary classification loss $\mathcal{L}^{\mathbf{D}_h}$.
Moreover, $s_{(h,o)}^\mathbf{D}$ will be used as the \textbf{final} interactiveness score in inference (Sec.~\ref{sec:test}).
Notably, we can only obtain the \textbf{instance-level} binary labels from instance-level HOI labels, but not the part-level binary labels. 

\subsubsection{Interactiveness Consistency}
\label{sec:d_ca}
With the help of part interactiveness, we can learn deeper interactiveness knowledge via the supervision of \textit{hierarchical consistency}.
In detail, a person is interactive if and only if at least one of the body parts is interacting with the object, and is not interactive if and only if none of the parts are interactive. 
This simple but robust relationship is an important clue for us to mine deeper discriminative information for complex and diverse HOIs. For example, in Fig.~\ref{Fig:local_ex}, Person\ 1 is riding the horse, which can be inferred by part features because his hands are holding the rope and feet are stepping on the pedals. In contrast, Person\ 2 and Person\ 3 are not interacting with the horse because none of their body parts are interacting with the horse.

Theoretically, instance interactiveness is equal to the result of \textit{OR} operation between all part interactivenesses. 
In practice, we can use the max-pooling to implement \textit{OR} operation. This is also in line with the MIL paradigm~\cite{Maron1998A}.
Thus, for a human-object pair, our prediction should obey:
\begin{align}
    p^{\mathbf{D}}_{(h,o)}=p^{\mathbf{D}}_{(hp,o)}=max (p^{\mathbf{D}}_{(p_i,o)}), 
    \label{eq:d_prob}   
\end{align}
where $p^{\mathbf{D}}_{(p_i,o)}(1 \leq i \leq 10)$ indicates the predicted interactiveness probability for the $i$-th body part. 
And $p^{\mathbf{D}}_{(hp,o)}$ is the instance interactiveness probability aggregated from part interactivenesses. 
We also use $p^{\mathbf{D}}_{(hp,o)}$ to generate another binary classification loss $\mathcal{L}^{\mathbf{D}_{hp}}$.
Meanwhile, $p^{\mathbf{D}}_{(h,o)}$ is the instance interactiveness probability from the instance binary classifier (Sec.~\ref{sec:binary-classifier}).
$max(\cdot)$ means max pooling operation.
In implementation, we use the predicted interactiveness scores to construct the \textbf{consistency loss}:
\begin{eqnarray}
    \mathcal{L}^{\mathbf{D}_c}=\mathbf{MSE}(s^{\mathbf{D}}_{(h,o)},max (s^{\mathbf{D}}_{(p_i,o)})),
\label{eq:loss_p_cons}
\end{eqnarray}
where MSE is the Mean Square Error, $s^{\mathbf{D}}_{(h,o)}$ means the instance interactiveness score and $s^{\mathbf{D}}_{(p_i,o)})$ indicates the $i$-th part interactiveness score. 
Consistency loss can avoid the conflicts between the knowledge from two different levels and bring stronger supervised guidance.
Overall, the loss of interactiveness discriminator $\mathbf{D}$ can be expressed as:
\begin{eqnarray}
    \mathcal{L}^{\mathbf{D}} = \mathcal{L}^{\mathbf{D}_h} + \mathcal{L}^{\mathbf{D}_{hp}}+
    \mathcal{L}^{\mathbf{D}_c},
\label{eq:loss_p_all}
\end{eqnarray}

\begin{figure}[!ht]
	\begin{center}
		\includegraphics[width=0.45\textwidth]{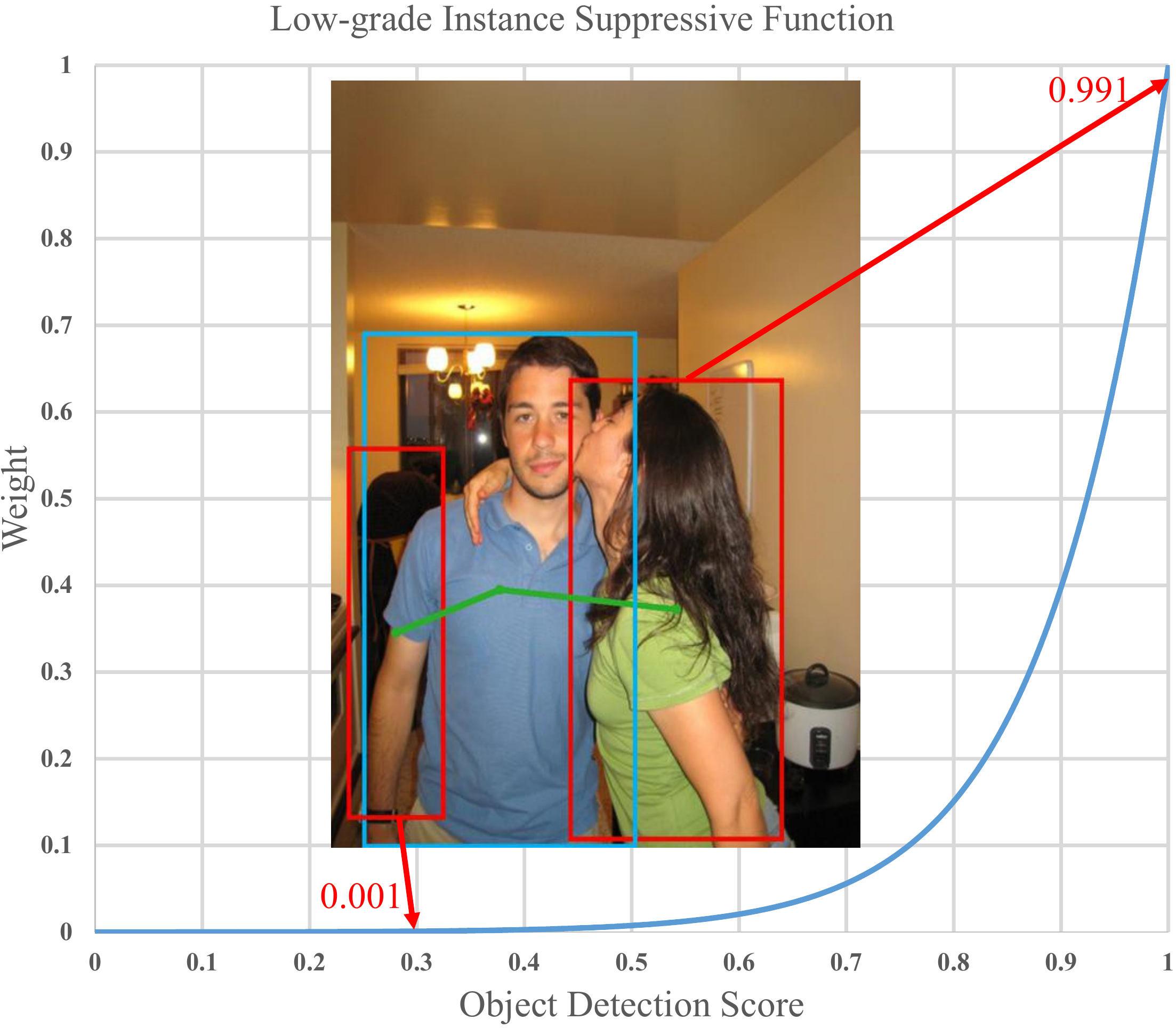}
	\end{center}
	\caption{The illustration of $\mathcal{P}(\cdot)$ within Low-grade Suppressive Function. Its input is object detection score. High-grade detected objects will be emphasized and distinguished with low-grade ones. In addition, $\mathcal{P}(0)=5.15E-05$ and $\mathcal{P}(1)=9.99E-01$. }
	\label{Figure:logictic}
	\vspace{-0.3cm}
\end{figure}

\subsubsection{Low-Grade Instance Suppressive Function}
\label{sec:lis}
Besides, we propose a Low-grade Suppressive Function (LIS) that can enhance the differentiation between high and low grade object detections. 
Given a HOI graph $\mathcal{G}$ with all possible edges, $\mathbf{D}$ will evaluate the interactiveness of pair $(v_h, v_o)$ based on learned knowledge, and gives a confidence $s^{D}_{(h,o)}$. Here, we further use LIS to modulate this confidence according to the human/object detection quality:
\begin{eqnarray}
    s'^{\mathbf{D}}_{(h,o)} = s^{\mathbf{D}}_{(h,o)} \ast L(s_h, s_o),
\label{eq:d_score}
\end{eqnarray}
where $L(s_h, s_o)$ is a novel weight function named Low-grade Instance Suppressive Function (LIS). It takes the human and object detection scores $s_h,s_o$ as inputs:
\begin{eqnarray}
    L(s_h, s_o) = \mathcal{P}(s_h) \ast \mathcal{P}(s_o),
\label{eq:lis}
\end{eqnarray}
where
\begin{eqnarray}
    \mathcal{P}(x) = \frac{T}{1+e^{(k-wx)}},
\label{eq:D_sp_func}
\end{eqnarray}
$\mathcal{P}(\cdot)$ is a part of the logistic function, the value of $\mathbf{T}$, $\mathbf{k}$ and $\mathbf{w}$ are determined by a data-driven manner. Fig.~\ref{Figure:logictic} depicts the curve of $\mathcal{P}(\cdot)$ whose domain definition is $(0,1)$. A bounding box would have low weight till its score is higher than a threshold. Previous works~\cite{gao2018ican,Gkioxari2017Detecting} often directly multiply detection scores by the final classification score. But they cannot notably emphasize the differentiation between high quality and inaccurate detection results. In contrast, LIS has the ability to enhance this differentiation as shown in Fig.~\ref{Figure:logictic}. 

\subsection{Inference with Non-Interaction Suppression}
\label{sec:test}
After the interactiveness learning,  we utilize $\mathbf{D}$ to suppress the non-interactive pair candidates in \textbf{testing}, i.e. Non-Interaction Suppression (NIS). 
The inference process is based on the tree structure as shown in Fig.~\ref{Figure:graph-refine}. Detected instances in the test set will be paired exhaustively, so a dense graph $\mathcal{G}$ of humans and objects is generated. 
First, we employ $\mathbf{D}$ to compute the interactiveness confidences of all edges. Next, we suppress the edges that meet NIS conditions, i.e. interactiveness confidence is smaller than a certain threshold $\alpha$.

Through NIS, we can convert $\mathcal{G}$ to $\mathcal{G}^{'}$ where $\mathcal{G}^{'}$ denotes the approximate sparse HOI graph. 
The HOI classification score vector $\mathcal{S}^{\mathbf{C}}_{(h, o)}$ of $(v_h, v_o)$ from $\mathbf{C}$ is denoted as:
\begin{eqnarray}
    \mathcal{S}^{\mathbf{C}}_{(h,o)} = \mathcal{F}_{{\mathbf{C}}}[\Gamma^{'}; \mathcal{G}^{'}(v_h, v_o)],
\label{eq:C_sparse_score}
\end{eqnarray}
where $\Gamma^{'}$ are input features. The final HOI score vector of a pair $(v_h, v_o)$ can be obtained by:
\begin{eqnarray}
\label{eq:final-score}
    \mathcal{S}_{(h,o)} = \mathcal{S}^{\mathbf{C}}_{(h,o)} \ast s'^{\mathbf{D}}_{(h,o)}.
\label{eq:pair_final_score}
\end{eqnarray}
Here we multiply interactiveness score $s'^{\mathbf{D}}_{(h,o)}$ from $\mathbf{D}$ (Eq.~\ref{eq:d_score}) by the output score of $\mathbf{C}$ (Eq.~\ref{eq:C_sparse_score}).

\section{Experiments}
\label{sec:experiment}
In this section, we first introduce the datasets and metrics adopted and then give the implementation details of our framework. Then we introduce two mode settings in the experiment, namely, \emph{default joint learning mode} and  \emph{transfer learning mode}. Next, we report our HOI detection results qualitatively and quantitatively compared with state-of-the-art approaches. Finally, we conduct ablation studies to validate the components in our framework. 

\subsection{Datasets and Metrics}
\label{sec:dataset}
We mainly adopt three HOI datasets: HICO-DET~\cite{hicodet}, V-COCO~\cite{vcoco} and HAKE~\cite{hake}.

\noindent{\bf HICO-DET~\cite{hicodet}} includes 47,776 images (38,118 in train set and 9,658 in test set), 600 HOI categories on 80 object categories (same with ~\cite{coco}) and 117 verbs, and provides more than 150k annotated human-object pairs.

\noindent{\bf V-COCO~\cite{vcoco}} provides 10,346 images (2,533 for training, 2,867 for validating and 4,946 for testing) and 16,199 person instances. Each person has annotations for 29 action categories (five of them have no paired object).
The objects are divided into two types: ``object'' and ``instrument''.

\noindent{\bf HAKE~\cite{hake}} provides 118K+ images, which include 285K human instances, 250K interacted objects, and 724K HOI pairs with human body part states~\cite{lu2018beyond}. The abundance of HOI samples can help our model achieve better performance on interactiveness discrimination.

\noindent{\bf PaStaNet-HOI~\cite{hake}}
To better evaluate our method, we re-split HAKE~\cite{hake} and construct a much larger benchmark: \textbf{PaStaNet-HOI}. It provides 110K+ images (77,260 images in train set, 11,298 images in validation set,  and 22,156 images in test set). Compared with HICO-DET~\cite{hicodet}, PaStaNet-HOI dataset has much larger train and test sets. The interaction categories are similar to the settings of HICO-DET~\cite{hicodet}, but we exclude the 80 ``non-interaction'' categories and only define 520 HOI categories. This can help to alleviate the \textit{annotation missing problem} in HICO-DET~\cite{hicodet}.

\noindent{\bf Metrics.} We follow the settings adopted in~\cite{hicodet}, i.e., a prediction is a true positive only when the human and object bounding boxes both have IoU larger than 0.5 with reference to ground truth, and the HOI classification result is accurate. The role mean average precision~\cite{vcoco} is used to measure the performance. Additionally, we measure the interactiveness detection in a similar setting. The only difference is that HOI classification is multi-label while the interactiveness classification is binary.

\subsection{Implementation Details}
We employ a Faster R-CNN~\cite{faster-rcnn} with ResNet-50~\cite{resnet} as $\mathbf{R}$ and keep it frozen.
$\mathbf{C}$ consists of three streams similar to~\cite{hicodet,gao2018ican}, extracting features $\mathbf{\Gamma^{'}}$ from instance appearance, spatial location as well as context. 
Within human/object stream, a residual block~\cite{resnet} with global average pooling and two 1024 sized FCs are used. 
Relatively, the spatial stream is composed of two convolutional layers with max-pooling, and two 1024 sized FCs. 
Following~\cite{hicodet,gao2018ican}, we use the late fusion strategy in $\mathbf{C}$. 

For the interactiveness network $\mathbf{D}$, the human stream, object stream, and spatial-pose stream are set the same as in $\mathbf{C}$. 
For part stream, after pooling, ten part features are first respectively concatenated with the instance-level features from the other three streams and then passed through two 1024 sized FCs to perform interactiveness discrimination. Whereafter, ten part interactiveness probabilities are generated. 
Max pooling is then imposed on them to reason out the aggregated instance interactiveness prediction.
Meanwhile, ten part features are multiplied by their corresponding interactiveness probabilities for re-weighting. 
Then, all ten body part features, together with other instance-level features are concatenated again and passed through two 1024 sized FCs to make instance interactiveness prediction. 
Finally, the consistency loss between two levels is constructed as the objective.

For a fair comparison, we adopt the object detection results and COCO~\cite{coco} pre-trained weights from~\cite{gao2018ican} which are provided by authors.
Since NIS and LIS can suppress non-interactive pairs, we set detection confidence thresholds lower than~\cite{gao2018ican}, i.e. 0.6 for humans and 0.4 for objects. 
The image-centric training strategy~\cite{faster-rcnn} is also applied. In other words, pair candidates from one image make up the mini-batch. 
Cross-validation is used to determine the hyper-parameters.
We adopt SGD with cosine decay restart policy and set an initial learning rate as 1e-4, momentum as 0.9. 
To handle the data bias, we control the ratio of positive and negative pairs for each image as 1:4.
We jointly train the framework for 25 epochs. In LIS mentioned in Eq.~\ref{eq:D_sp_func}, we set $T=8.4,k=12.0,w=10.0$.
In testing, the interactiveness threshold $\alpha$ in NIS is set as 0.1. All experiments are conducted on a single Nvidia Titan X GPU.

\subsection{Default Joint Learning}
In \emph{default joint learning mode},
HOI classifier $\mathbf{C}$ is trained together with $\mathbf{D}$.
By adding a supervisor $\mathbf{D}$, our framework works in an unconventional training mode. 
To be specific, the framework is trained with hierarchical classification tasks, i.e. explicit interactiveness discrimination and HOI classification.
The overall loss of the proposed method can be expressed as:
\begin{eqnarray}
    \mathcal{L} = \mathcal{L}^{\mathbf{C}} + \mathcal{L}^{\mathbf{D}},
\label{eq:loss}
\end{eqnarray}
where $\mathcal{L}^{\mathbf{C}}$ denotes the HOI classification cross entropy loss, $\mathcal{L}^{\mathbf{D}}$ is the binary classification cross entropy loss (Eq. \ref{eq:loss_p_all}).

Different from one-stage methods, additional interactiveness discrimination enforces the model to learn interactiveness knowledge, which can bring more powerful constraints. Namely, when a pair is predicted as specific HOIs such as ``cut cake'', $\mathbf{D}$ must give the prediction ``interactive'' simultaneously. 
Experiment results (Sec.~\ref{sec:ablation}) prove that interactiveness knowledge learning can effectively refine training and improve performance. 

An additional benefit of the default joint learning mode is that, if cooperated with the multi-stream model $\mathbf{C}$, $\mathbf{D}$ can share the weights of convolutional blocks with the ones in $\mathbf{C}$. To be more specific, blocks $H^{\mathbf{D}}$ and $O^{\mathbf{D}}$ can share weights with $H^C$ and $O^C$ in the joint training. This weights sharing strategy can guarantee information sharing and better optimization of $\mathbf{D}$ and $\mathbf{C}$ in the multi-task training. 

The framework in default joint learning mode is called ``$\mathbf{R}\mathbf{C}\mathbf{D}$'' in the following, where ``$\mathbf{R}$'' ``$\mathbf{C}$'' ``$\mathbf{D}$'' represents the representation network, the classification network, and the interactiveness network respectively.

\subsection{Transfer Learning}
As aforementioned, $\mathbf{D}$ only needs binary labels that can be easily converted from HOI labels and are beyond HOI classes. 
On account of the generalization ability of interactiveness, $\mathbf{D}$ can be trained and transferred across datasets.
Therefore, in \emph{transfer learning mode}, $\mathbf{D}$ is used as a transferable knowledge learner to learn interactiveness from multi-dataset, and then be applied to each of them respectively. To be more specific, we train $\mathbf{C}$ and $\mathbf{D}$ \textbf{separately} instead of jointly.
In the transfer learning mode, the benefit of sharing weight is \textit{absent}. However, the independent training makes it possible for $\mathbf{D}$ to learn interactiveness knowledge across datasets, i.e., \textit{flexibility} and \textit{reusability}. 
Furthermore, the integration of datasets boosts the performance of interactiveness learning. So we can use the cross-trained $\mathbf{D}$ to operate a more powerful NIS in inference.

The mode settings of transfer learning are shown in Tab.~\ref{tab:mode}. The train and test sets of $\mathbf{C}$ are kept the same as default mode, while the train set of $\mathbf{D}$ is adjusted. 
For transfer learning mode $\mathbf{R}\mathbf{C}\mathbf{D}_{i}$, increasingly larger train sets are marked by $i$ from $i=1$ to $i=3$, i.e., \textbf{1)} V-COCO or HICO-DET, \textbf{2)} V-COCO and HICO-DET, \textbf{3)} HAKE. 
That said, $\mathbf{D}$ will be trained with more and more samples. Thus we can analyze the effect of NIS with different scales of interactiveness knowledge.

Meanwhile, different from $\mathbf{D}$, $\mathbf{C}$ must be trained on a single dataset once a time considering the variety of HOI category settings in different datasets. Therefore, knowledge of the specific HOIs is difficult to transfer. We will compare and evaluate the transfer abilities of interactiveness knowledge and HOI knowledge in Sec.~\ref{sec:ablation}.

\begin{table}
\centering
\resizebox{0.48\textwidth}{!}{
\begin{tabular}{l c c c}
\hline
Test Set &  Method   & $\mathbf{D}$-Train Set & $\mathbf{C}$-Train Set   \\
\hline
\hline
\multirow{4}{2.3cm}{HICO-DET}    
& $\mathbf{R}\mathbf{C}\mathbf{D}$      & HICO-DET & HICO-DET\\
~   & $\mathbf{R}\mathbf{C}\mathbf{D}_{1}$      & V-COCO & HICO-DET\\
~   & $\mathbf{R}\mathbf{C}\mathbf{D}_{2}$  & HICO-DET, V-COCO  & HICO-DET  \\
~   & $\mathbf{R}\mathbf{C}\mathbf{D}_{3}$  & HAKE  & HICO-DET\\
\hline
\multirow{4}{2.3cm}{V-COCO}  
& $\mathbf{R}\mathbf{C}\mathbf{D}$        & V-COCO & V-COCO   \\
~       & $\mathbf{R}\mathbf{C}\mathbf{D}_{1}$        & HICO-DET & V-COCO   \\
~       & $\mathbf{R}\mathbf{C}\mathbf{D}_{2}$        & HICO-DET, V-COCO & V-COCO    \\
~       & $\mathbf{R}\mathbf{C}\mathbf{D}_{3}$        & HAKE & V-COCO    \\
\hline
\multirow{1}{2.3cm}{PaStaNet-HOI}  
~       & $\mathbf{R}\mathbf{C}\mathbf{D}$        & PaStaNet-HOI & PaStaNet-HOI   \\
\hline
\end{tabular}}
\caption{Mode settings. $\mathbf{R}\mathbf{C}\mathbf{D}$ is the \textit{default joint learning mode} while $\mathbf{R}\mathbf{C}\mathbf{D}_{i}$ is the \textit{transfer learning mode}. $\mathbf{R}\mathbf{C}\mathbf{D}_{i}$ means that $\mathbf{D}_{i}$ is trained with increasingly larger datasets marked by $i$ from $1$ to $3$.}
\label{tab:mode}
\vspace{-0.3cm}
\end{table}

\subsection{Results and Comparisons}
\label{sec:score}
We compare our method with five state-of-the-art HOI detection methods~\cite{hicodet,Gkioxari2017Detecting,qi2018learning,gao2018ican,peyre2018detecting} on HICO-DET, and four methods~\cite{vcoco,Gkioxari2017Detecting,qi2018learning,gao2018ican} on V-COCO.
The result is evaluated with mAP. 
For HICO-DET, we follow~\cite{hicodet}: Full (600 HOIs), Rare (138 HOIs), Non-Rare (462 HOIs) in Default and Known Object modes. For V-COCO, we evaluate $AP_{role}$ (24 actions with roles) on Scenario 1. For PaStaNet-HOI, we evaluate mAP following the Default mode of~\cite{hicodet} for 520 HOIs. More details can be found in~\cite{hicodet,vcoco}. 
                          
\begin{table}
\centering
\resizebox{0.48\textwidth}{!}{
\begin{tabular}{l  c  c  c  c  c  c  }
\hline
         & \multicolumn{3}{c}{Default}  &\multicolumn{3}{c}{Known Object} \\
Method         & Full & Rare & Non-Rare  & Full & Rare & Non-Rare \\
\hline
\hline
HO-RCNN~\cite{hicodet}       & 7.81  & 5.37     & 8.54     & 10.41   & 8.94   & 10.85\\
InteractNet~\cite{Gkioxari2017Detecting} & 9.94  & 7.16     & 10.77     & -   & -   & -\\
GPNN~\cite{qi2018learning}   & 13.11  & 9.34     & 14.23     & -   & -   & -\\
iCAN~\cite{gao2018ican}     & 14.84  & 10.45     & 16.15     & 16.26   & 11.33   & 17.73\\
Peyre et al.~\cite{peyre2018detecting}&
19.40 & 14.60 & 20.90 & - & - & -\\
\hline
$\mathbf{R}\mathbf{C}\mathbf{D}$ &17.84 & 13.08 & 18.78 & 20.58 & 16.19 & 21.45\\
\hline
$\mathbf{R}\mathbf{C}\mathbf{D}_{1}$               & 17.49 & 12.23 & 18.53 & 20.28 & 15.25 & 21.27\\
$\mathbf{R}\mathbf{C}\mathbf{D}_{2}$               & 18.43 & 13.93 & 19.32 & 21.10 & 16.56 & 22.00\\
$\mathbf{R}\mathbf{C}\mathbf{D}_{3}$               & \textbf{20.93} & \textbf{18.95} & \textbf{21.32} & \textbf{23.02} & \textbf{20.96} & \textbf{23.42}\\
$\mathbf{R}$+iCAN~\cite{gao2018ican}+$\mathbf{D}_3$ & 17.58 & 13.75 & 18.33 & 19.13 & 15.06 & 19.94\\
\hline
\end{tabular}}
\caption{Results comparison on HICO-DET~\cite{hicodet}.} 
\label{tab:hico-det}
\end{table}

\begin{table}
\centering
\resizebox{0.32\textwidth}{!}{
\begin{tabular}{l c }
\hline
Method         &$AP_{role}$\\
\hline
\hline
Gupta et al.~\cite{vcoco}                           & 31.8 \\
InteractNet~\cite{Gkioxari2017Detecting}           & 40.0 \\
GPNN~\cite{qi2018learning}                         & 44.0 \\
iCAN w/ late(early)~\cite{gao2018ican}             & 44.7 (45.3) \\
\hline
 $\mathbf{R}\mathbf{C}\mathbf{D}$                         & 48.4 \\
\hline
$\mathbf{R}\mathbf{C}\mathbf{D}_1$                         & 48.5 \\
$\mathbf{R}\mathbf{C}\mathbf{D}_2$                         & 48.7 \\
$\mathbf{R}\mathbf{C}\mathbf{D}_3$                         & \textbf{49.1} \\
$\mathbf{R}$+iCAN w/ late(early)~\cite{gao2018ican}+ $\mathbf{D}_{3}$            & 45.8(46.1) \\
\hline
\end{tabular}}
\caption{Results comparison on V-COCO~\cite{vcoco}.} 
\label{tab:vcoco}
\vspace{-0.3cm}
\end{table}

\subsubsection{Results Analysis}
Results are shown in Tab.~\ref{tab:hico-det}, Tab.~\ref{tab:vcoco}, and Tab.~\ref{tab:hake} respectively.

\noindent {\bf Default Joint Learning.} The $\mathbf{RC}$ modules adopt the similar model design, object detection, and backbone with previous methods~\cite{hicodet,gao2018ican}.
However, with the joint learning and NIS, our $\mathbf{RCD}$ directly outperforms \cite{hicodet,gao2018ican} with 10.03 and 3.00 mAP on HICO-DET. On V-COCO and PaStaNet-HOI respectively, $\mathbf{RCD}$ also achieves 3.10 mAP and 4.38 mAP improvements compared with iCAN~\cite{gao2018ican}. This greatly verifies the efficacy of hierarchical interactiveness learning.

\noindent {\bf Transfer Learning.} 
In transfer learning, with the size of $\mathbf{D}$ train set increases, the performance is also improved accordingly.
$\mathbf{R}\mathbf{C}\mathbf{D}_3$ presents the greatest improvement and achieves the state-of-the-art performance, i.e., \textbf{20.93}, \textbf{18.95}, \textbf{21.32} mAP on three Default sets on HICO-DET, and \textbf{49.1} mAP on V-COCO. It surpasses the recent state-of-the-art~\cite{peyre2018detecting} by \textbf{1.53}, \textbf{4.35}, and \textbf{0.42} mAP on three Default sets on HICO-DET. Compared with mode $\mathbf{R}\mathbf{C}\mathbf{D}_1$, $\mathbf{R}\mathbf{C}\mathbf{D}_3$ gains an improvement of \textbf{3.44} and \textbf{2.74} mAP on Default and Known Object Full sets on HICO-DET, and an increase of \textbf{6.72} and \textbf{5.71} mAP on Rare sets.
Noticeably, as the generalization ability of interactiveness is beyond HOI category settings, information scarcity and learning difficulty of rare HOI categories is alleviated. So the performance difference between rare and non-rare HOI categories is accordingly reduced. 
Our method achieves excellent performance on rare HOIs, the Rare set performance of $\mathbf{R}\mathbf{C}\mathbf{D}_3$ (18.95 mAP) is even beyond the Non-Rare performances of  $\mathbf{R}\mathbf{C}\mathbf{D}$ (18.78 mAP) and previous methods~\cite{gao2018ican} (16.15 mAP).
This strongly proves the transferability of interactiveness. 

On the other hand, we also apply NIS to iCAN~\cite{gao2018ican} and compare the testing results with the original ones presented in \cite{gao2018ican}. Results show that NIS also brings an increase of \textbf{2.74} and \textbf{2.87} mAP on Default and Known Object Full sets on HICO-DET, \textbf{1.1} and \textbf{0.8} mAP on V-COCO, and \textbf{2.13} mAP on PaStaNet-HOI. This again proves the versatility and effectiveness of the proposed NIS.

\begin{table}[!t]
    \centering
    \begin{minipage}[t]{0.38\linewidth}
    \centering
    \resizebox{\textwidth}{!}{\setlength{\tabcolsep}{0.6mm}{
      \begin{tabular}{l  c  c  c}
        \hline
        Method & mAP\\
        \hline
        \hline
        iCAN~\cite{gao2018ican}&11.00\\
        \hline
        $\mathbf{R}\mathbf{C}\mathbf{D}$ &\textbf{15.38}\\
        $\mathbf{R}$+iCAN~\cite{gao2018ican}+$\mathbf{D}_3$& 13.13\\
        \hline
        \end{tabular}}}
        \caption{Results on PaStaNet-HOI.} 
        \label{tab:hake}
    \end{minipage}
    \begin{minipage}[t]{0.45\linewidth}
      \centering
      \resizebox{\textwidth}{!}{\setlength{\tabcolsep}{0.6mm}{
        \begin{tabular}{l  l c }
        \hline
        Test Set      & Method                  & mAP\\
        \hline
        \hline
        HICO-DET & $\mathbf{R}\mathbf{C}\mathbf{D}$ & 14.35\\
        HICO-DET & $\mathbf{R}\mathbf{C}\mathbf{D}_3$ & 16.42\\
        \hline
        PaStaNet-HOI & $\mathbf{R}\mathbf{C}\mathbf{D}$ & 19.29\\
        \hline
        \end{tabular}}}
      \caption{\textbf{Interactiveness detection} result.}
      \label{tab:binary}
    \end{minipage}
\vspace{-0.1cm}
\end{table}

\begin{table}
\centering
\resizebox{0.27\textwidth}{!}{
\begin{tabular}{l  c c c}
\hline
Test Set      & Method                  & Reduction\\
\hline
\hline
HICO-DET & $\mathbf{R}\mathbf{C}\mathbf{D}$ & -68.73\%\\
\hline
\multirow{3}{1.8cm}{HICO-DET}  & $\mathbf{R}\mathbf{C}\mathbf{D}_1$      & -67.75\% \\
~   & $\mathbf{R}\mathbf{C}\mathbf{D}_2$                    & -74.52\%\\
~   & $\mathbf{R}\mathbf{C}\mathbf{D}_3$                & -85.93\%\\
\hline
\hline
V-COCO & $\mathbf{R}\mathbf{C}\mathbf{D}$ & -50.75\%\\
\hline
\multirow{3}{1.8cm}{V-COCO} & $\mathbf{R}\mathbf{C}\mathbf{D}_1$                & -59.32\%\\
~   & $\mathbf{R}\mathbf{C}\mathbf{D}_2$                    & -61.55\%\\
~   & $\mathbf{R}\mathbf{C}\mathbf{D}_3$                & -72.19\%\\
\hline
\hline
PaStaNet-HOI & $\mathbf{R}\mathbf{C}\mathbf{D}$ & -86.20\%\\
\hline
\end{tabular}}
\caption{Non-interactive pairs reduction after performing NIS.}
\label{tab:transfer}
\end{table}

Since NIS implicitly evaluates the performance of our interactiveness network, we also \textit{explicitly} measure the interactiveness detection performance via role mean average precision~\cite{vcoco} (Tab.~\ref{tab:binary}). $\mathbf{R}\mathbf{C}\mathbf{D}$ achieves 14.35 and 19.29 mAP respectively on HICO-DET~\cite{hicodet} and PaStaNet-HOI~\cite{hake}. Benefit from a larger training set,  $\mathbf{R}\mathbf{C}\mathbf{D}_3$ outperforms  $\mathbf{R}\mathbf{C}\mathbf{D}$ by 2.07 mAP on HICO-DET~\cite{hicodet}.

To further study the relationship between the performance improvement and the training sample scale of $\mathbf{D}$, we analyze the non-interactive pairs reduction after employing NIS (Tab.~\ref{tab:transfer}).
As we cannot annotate all non-interactive pairs in the test set, we compute the ratio according to the assigned binary labels. 
Especially, HICO-DET has 80 ``no-interaction'' HOIs corresponding to 80 objects. They are ``naturally'' non-interactive but ``positive'' for the HICO-DET setting. Therefore, when the NIS condition is satisfied (NIS score is lower than the threshold), we do not discard the pairs belonging to these 80 ``no-interaction'' HOIs and only discard the pairs of other 520 normal HOIs.
With interactiveness transferred from more data, $\mathbf{R}\mathbf{C}\mathbf{D}_3$
achieves the best suppressive effect and discards \textbf{85.93\%} and \textbf{72.19\%} non-interactive pairs respectively on two datasets, thus bringing more performance gains.
Meanwhile, $\mathbf{R}\mathbf{C}\mathbf{D}_1$, $\mathbf{R}\mathbf{C}\mathbf{D}_2$ also perform well and suppresses a certain amount of non-interactive pair candidates. 
This indicates the great potential of the interactiveness network. Namely, with more training data, $\mathbf{D}$ is expected to boost the performance further.
Moreover, since HICO-DET train set (38K) is much larger than V-COCO train set (2.5K), the default mode $\mathbf{R}\mathbf{C}\mathbf{D}$ achieves larger improvement than $\mathbf{R}\mathbf{C}\mathbf{D}_1$ on HICO-DET (Tab.~\ref{tab:hico-det}).
And the situation between $\mathbf{R}\mathbf{C}\mathbf{D}$ and $\mathbf{R}\mathbf{C}\mathbf{D}_1$ on V-COCO is opposite.
On PaStaNet-HOI, we also find an effective non-interaction pairs reduction (86.20\%).

\subsubsection{Visualized Results}
\begin{figure}[!ht]
	\begin{center}
		\includegraphics[width=0.48\textwidth]{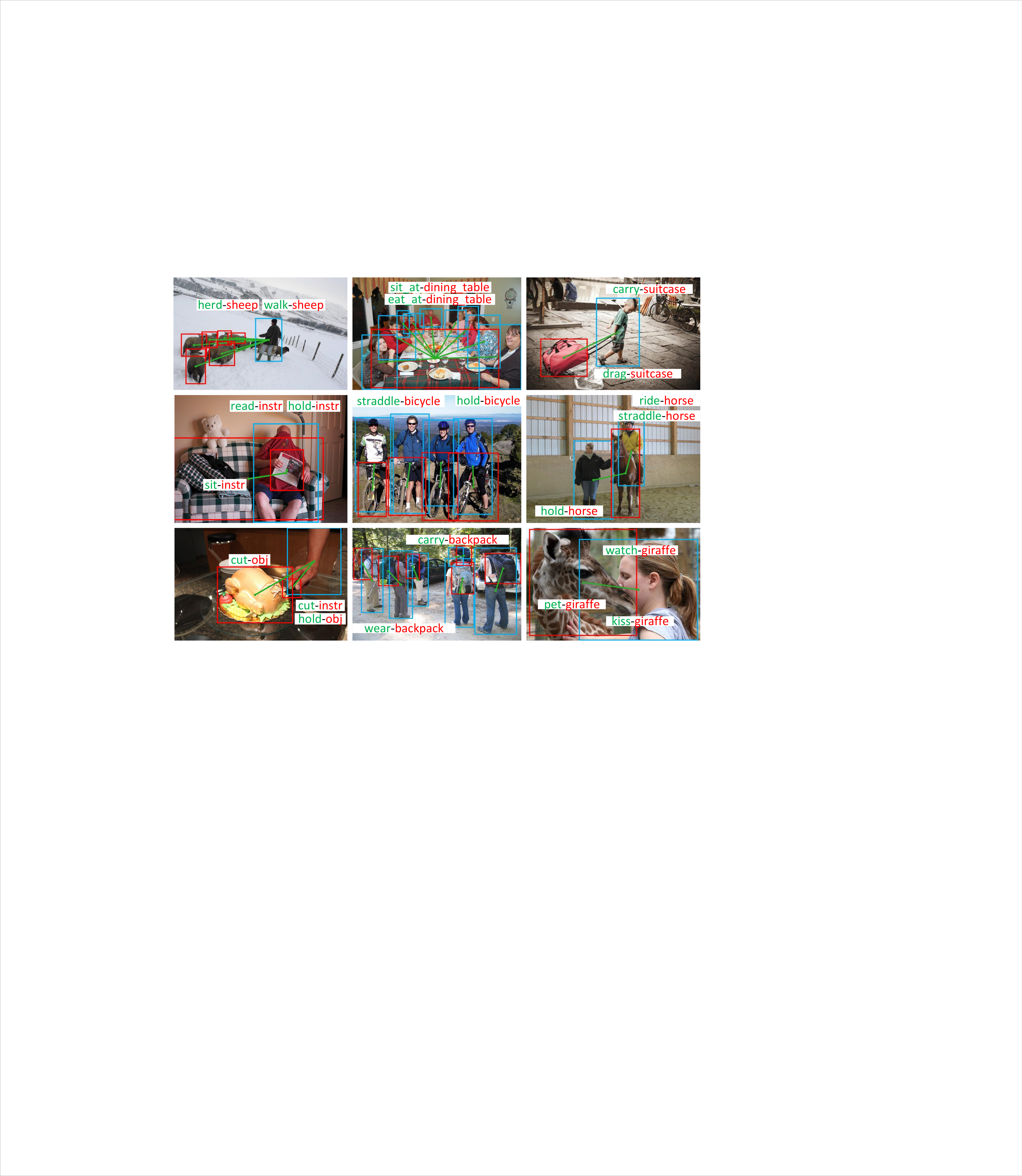}
	\end{center}
	\caption{Visualization of sample HOI detections. Subjects and objects are represented with blue and red bounding boxes. While interactions are marked by green lines linking the box centers.}
	\label{Figure:vis}
	\vspace{-0.3cm}
\end{figure}

Representative predictions are shown in Fig.~\ref{Figure:vis}. We find that our model is capable of detecting various complicated HOIs, such as multiple interactions within one pair, one person performing multiple interactions with different objects, one object interacted with multiple persons, multiple persons performing different interactions with multiple objects.

\begin{figure}[!ht]
	\begin{center}
		\includegraphics[width=0.5\textwidth]{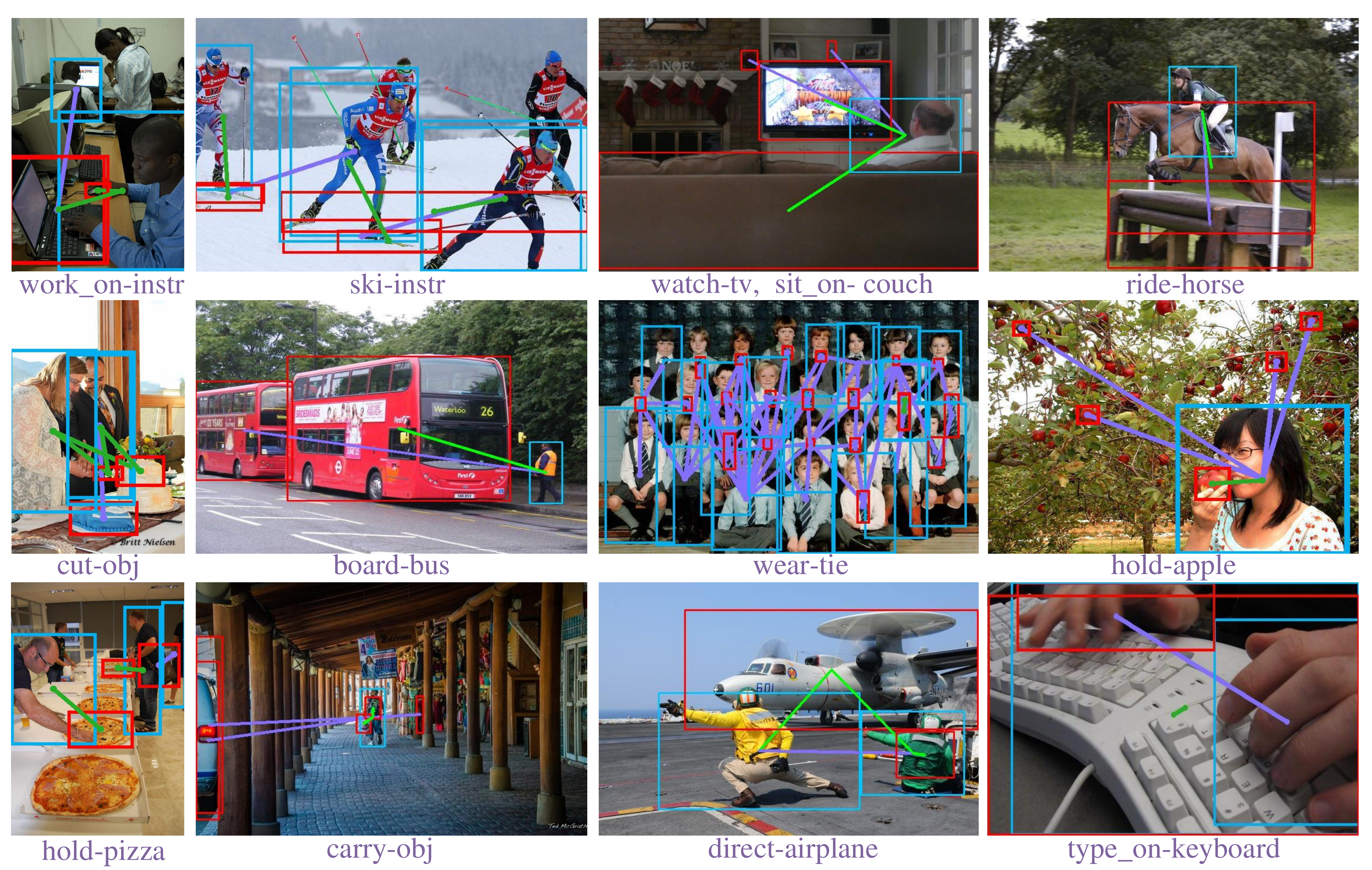}
	\end{center}
	\caption{Visualized effects of NIS. Green lines mean accurate HOIs, while purple lines mean non-interactive pairs which are suppressed. Without NIS, $\mathbf{C}$ would generate false positive predictions for these non-interactive pairs in one-stage inference, which are shown by the purple texts below the images.
	Even some extremely hard scenarios can be discovered and suppressed, such as mis-groupings between person and object close to each other, person and object in clutter scene.}
	\label{Figure:nis_vis}
	\vspace{-0.3cm}
\end{figure}

\begin{figure*}[!ht]
	\begin{center}
		\includegraphics[width=0.95\textwidth]{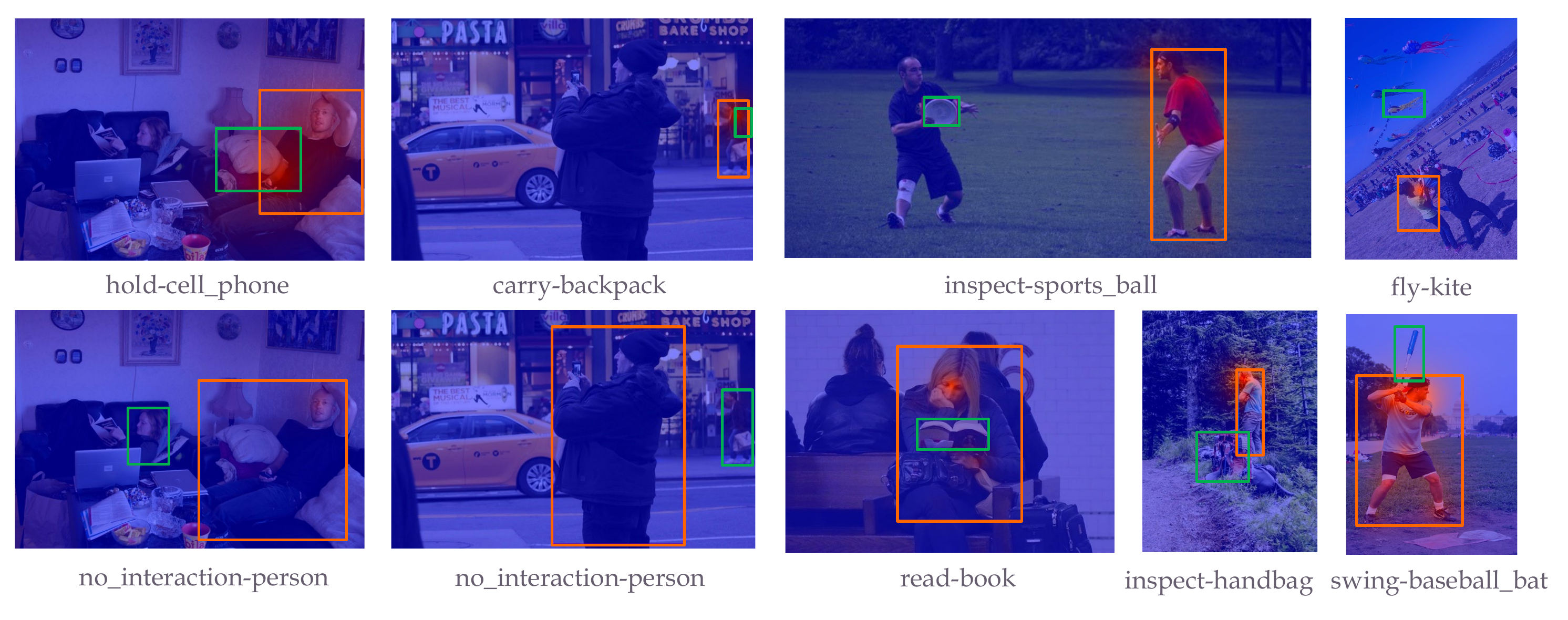}
	\end{center}
	\caption{The heatmaps of interactiveness attention based on $\mathbf{R}\mathbf{C}\mathbf{D}_3$ on HICO-DET and $\mathbf{R}\mathbf{C}\mathbf{D}$ on PaStaNet-HOI. The pixels with higher interactiveness probabilities are presented with a brighter red color. We can find that part-level interactiveness knowledge localizes the most informative parts effectively. With the interactiveness heatmaps, we can take further insight into the model.}
	\label{Figure:vis_heatmap}
	\vspace{-0.3cm}
\end{figure*}

Fig.~\ref{Figure:nis_vis} shows the visualized effects of NIS. We can see that NIS effectively distinguishes the non-interactive pairs and suppresses them in extremely difficult scenarios, such as a person performing a confusing action and a crowd of people with ties. In the bottom-right corner, we show an even harder sample.
For the HOI ``type\_on keyboard'' between human and the keyboard, $\mathbf{C}$ wrongly predicts that this HOI is within two close hands.
However, $\mathbf{D}$ accurately figures out that two hands are non-interactive. These results prove that the one-stage method would yield many false positives without interactiveness and NIS. 

\subsubsection{Part Interactiveness Insight}
\label{sec:pattern}
As aforementioned, we use part attention strategy to strengthen the important parts in inference. 
With $\mathbf{R}\mathbf{C}\mathbf{D}_3$ (HICO-DET) and $\mathbf{R}\mathbf{C}\mathbf{D}$ (PaStaNet-HOI), we visualize and quantify the part interactiveness patterns for certain HOIs.

Fig.~\ref{Figure:vis_heatmap} illustrates the interactiveness attention heatmaps. Pixels with higher interactiveness probabilities are presented with a brighter red color. 
We find that part-level interactiveness knowledge localizes the most informative parts effectively. 
For example, for ``swing-baseball\_bat'' (row2, col5), hands and arms have obvious higher attention, while hip, thighs, and feet are not lightened. 
With the interactiveness heatmaps, we can take a further insight into the model. We first compare HOI and non-HOI cases (col1 and col2). For HOI cases, the most possible interactive parts are localized, while body parts all get low interactiveness inference scores for non-HOI cases. Furthermore, we find that the model can learn some functionality of different human body parts rather than simply highlight body parts \textit{physically close} to the object. 
For ``inspect-sports\_ball'' (row1, col3) and ``fly-kite'' (row1, col4), ``head'' and ``hands'' are respectively highlighted although 10 body parts are equally far away from the object. 
For ``inspect-handbag'' (row2, col4), ``head'' is highlighted instead of the closer ``feet'' and ``legs'', which verifies the effectiveness of our interactiveness network. Another similar case is ``read-book'' (row2, col3).

We also list the quantified results in Tab.~\ref{tab:pattern}. For a certain HOI, we first select all the related pairs on HICO-DET and then calculate the average interactiveness scores of each part respectively. For symmetrical parts (feet, thighs, upper arms, and hands), the model processes the left and right parts separately with different bounding boxes. 
For simplicity, we average the ``left'' and ``right'' scores for the same kind of parts.
Finally, for the six parts (feet, thighs, hip, upper arms, hands, and head), we use a minimax-scaler to rescale the six interactiveness scores to $[0,1]$ for each HOI, \textit{i.e.,} the highest interactiveness score is scaled to 1.0 while the lowest to 0.0.

\begin{table}
\centering
\resizebox{0.48\textwidth}{!}{
\begin{tabular}{l| c c c c c c}
\hline
HOI         &Feet & Thighs & Hip & Upper arms & Hands & Head \\
\hline
\hline
feed cat & 0.50 & 1.00 & 0.00 & 0.81 & 0.52 & 0.19\\
pet cat & 0.09 & 1.00 & 0.62 & 0.87 & 0.35 & 0.00\\
chase cat & 1.00 & 0.75 & 0.00 & 0.58 & 0.79 & 0.80\\
lie-on couch & 0.39 & 1.00 & 0.61 & 0.65 & 0.40 & 0.00\\
sit-on couch & 1.00 & 0.89 & 0.00 & 0.49 & 0.40 & 0.10\\
board airplane & 1.00 & 0.63 & 0.00 & 0.73 & 0.77 & 0.66\\
hold hair-dryer & 0.09 & 1.00 & 0.51 & 0.75 & 0.33 & 0.00\\
stand-under stop-sign & 0.08 & 1.00 & 0.00 & 0.85 & 0.43 & 0.21\\
\hline
\end{tabular}}
\caption{Body parts interactiveness attention pattern of $\mathbf{R}\mathbf{C}\mathbf{D}_3$ on HICO-DET~\cite{hicodet}. 
For a certain HOI, we first select all the related pairs on HICO-DET~\cite{hicodet} and then calculate the average interactiveness scores of each part respectively. Finally, we use a minimax-scaler to rescale the six interactiveness scores to $[0,1]$ for each HOI, i.e., the highest interactiveness score is scaled to 1.0 while the lowest to 0.0.}
\label{tab:pattern}
\vspace{-0.3cm}
\end{table}

The results in Tab.~\ref{tab:pattern} are insightful. 
For ``cat'', different actions have resulted in different part interactiveness patterns. ``feed cat'' is most related to thighs and upper arms, while ``chase cat'' is most related to feet, hands and head. 
This is consistent with common sense. 
When feeding a cat, a person may let the cat sit on thighs and use arms to give it food. Meanwhile, when chasing a cat, a person may move the feet quickly and try to catch it by hand.
Besides, there are also some inaccurate situations. Upper arms and hands are sometimes confused. For ``hold hair-dryer'', hands should be more interactive than shoulders, but reverse results are generated. 
For ``sit-on couch'', the attention of the hip is zero, which is unreasonable. 
Such biases are probably caused by the occlusion. In many training images, the hip is usually invisible and occluded by other objects. 

\subsection{Ablation Studies}
\label{sec:ablation}
In $\mathbf{R}\mathbf{C}\mathbf{D}$, we analyze the significance of Low-grade Instance Suppressive and Non-Interaction Suppression in inference (Tab.~\ref{tab:ablation}). We also analyze the design of the interactiveness network and the transferability of HOI Knowledge.

\noindent{\bf{Non-Interaction Suppression.}} NIS plays a key role to reduce the non-interactive pairs. We evaluate its impact by removing NIS during testing. 
In other words, we do not use NIS to discard the non-interactive pairs and directly use the $\mathcal{S}^{\mathbf{C}}_{(h,o)}$ (Eq.~\ref{eq:C_sparse_score}) as the final prediction. 
Consequently, the model shows an obvious performance degradation, which proves the importance of NIS.

\noindent{\bf{Low-grade Instance Suppressive. }} LIS suppresses the low-grade object detections and rewards the high-grade ones. By removing $L(s_h, s_o)$ in Eq.~\ref{eq:d_score}, we observe a degradation in Tab.~\ref{tab:ablation}. This suggests that LIS is capable of distinguishing the low-grade detections and improves the performance without using a more costly superior object detector. 

\noindent{\bf{NIS \& LIS.}} Without NIS and LIS both, our method only takes effect in the \textit{joint training}. As we can see in Tab.~\ref{tab:ablation}, performance degrades greatly but still outperforms iCAN~\cite{gao2018ican}, which indicates the enhancement brought by $\mathbf{D}$ in the hierarchical joint training.

\begin{table}
\centering
\resizebox{0.5\textwidth}{!}{
\begin{tabular}{l  c c c c}
\hline
&HICO-DET &V-COCO&PaStaNet-HOI \\
Method         & Default Full & $AP_{role}$ & mAP on Validation Set\\
\hline
\hline
iCAN~\cite{gao2018ican}& 14.84 &  45.3 & 20.45\\
$\mathbf{R}\mathbf{C}\mathbf{D}$ & 17.84 &  48.4&26.92\\
w/o NIS                 &15.86         &46.2&24.86\\
w/o LIS                 &16.35   &47.4& 26.33\\
w/o NIS \& LIS          &15.45    &45.8& 24.57\\
\hline
$\mathbf{R}\mathbf{C}\mathbf{D}_3$ ( $\mathbf{R}\mathbf{C}\mathbf{D}$ for PaStaNet-HOI)  & 20.93      & 49.1&26.92\\
w/o part stream & 18.52 & 48.7&25.60\\
w/o max-pooling & 18.85 & 48.7 &25.58 \\
w/o separate binary classifier &20.18 & 48.9 & 26.27\\ 
\hline
$\mathbf{R}\mathbf{C}_\mathbf{T}$ & 10.61  & 38.5 &-\\
\hline
\end{tabular}}
\caption{Results of ablation studies. Following~\cite{hicodet,qi2018learning,gao2018ican,peyre2018detecting}, we report the results on HICO-DET and V-COCO \textit{test} sets. On PaStaNet-HOI, the results on the \textit{validation} set are shown.}
\label{tab:ablation}
\vspace{-0.1cm}
\end{table}

\noindent{\bf{Model Design.}}
To verify the components of our model, we train $\mathbf{D}$ without the part stream, i.e., only with three conventional streams: human, object, and spatial-pose.
We also train our model without the max-pooling stream, i.e., w/o part-level interactiveness $s_{(hp,o)}^\mathbf{D}$ and consistency loss $\mathcal{L}^{\mathbf{D}_c}$.
Our model shows obvious degradation in these situations, which strongly verifies the effectiveness of the hierarchical interactiveness paradigm. 
Additionally, we adopt 10 classifiers for 10 parts. Alternatively, we can also use a shared part interactiveness classifier only. As illustrated in Tab.~\ref{tab:trade-off}, this modification also degrades the performance, despite slight improvements in computation properties (training/inference speed, model size, parameter).

\begin{table}
\centering
\resizebox{0.45\textwidth}{!}{
\begin{tabular}{l c c c c}
\hline
& Setting/Dataset & Separate& Shared \\
\hline
\hline
\multirow{2}{1.8cm}{Speed}  & training & 1.63 FPS & 1.85 FPS \\
~   & inference & 4.9 FPS & 5.5 FPS\\
\hline
\multirow{2}{1.8cm}{Size} & training & 10.4 GB & 9.9 GB\\
~   & inference & 7.3 GB & 7.0 GB\\
\hline
Parameter & - & 290 M & 141 M\\
\hline
\multirow{2}{1.8cm}{Performance} & HICO-DET~\cite{hicodet} & 20.93 mAP&20.18 mAP\\
~   & PaStaNet-HOI~\cite{hake} & 26.92 mAP& 26.27 mAP\\
\hline
\end{tabular}}
\caption{Performance and computation comparison on interactiveness network with separate/share interactiveness classifier for 10 body parts. As for the performance, following~\cite{hicodet,qi2018learning,gao2018ican,peyre2018detecting}, we report the results on HICO-DET \textit{test} sets. On PaStaNet-HOI, the results on the \textit{validation} set are shown.}
\label{tab:trade-off}
\vspace{-0.3cm}
\end{table}

\noindent{\bf{Transferability of HOI Knowledge.}} We also evaluate the transferability of HOIs. $\mathbf{R}\mathbf{C}_\mathbf{T}$ means $\mathbf{C}$ is trained on HICO-DET and tested on V-COCO, and vice versa.
Compared with iCAN~\cite{gao2018ican}, it shows a significant performance decrease of 4.23 and 6.80 mAP on two datasets. On the contrary, $\mathbf{D}$ shows satisfying performance across different datasets. This proves that interactiveness is more suitable and easier to transfer than HOI knowledge.

\section{Conclusion}
In this paper, we propose a novel method to learn and utilize the implicit interactiveness knowledge, which is general and beyond HOI categories. Thus, it can be transferred across datasets. We propose a hierarchical interactiveness paradigm to adopt both instance and part level interactivenesses. And a consistency learning task is further explored to improve the learning. 
With interactiveness knowledge, we exploit an interactiveness network to perform Non-interaction Suppression before HOI classification in inference. 
Extensive experiment results show the efficiency of learned interactiveness knowledge. By combining our method with existing detection models, we achieve state-of-the-art results on the HOI detection task.

\section*{Acknowledgment}
This work is supported in part by the National Key R\&D Program of China, No. 2017YFA0700800, National Natural Science Foundation of China under Grants 61772332, Shanghai Qi Zhi Institute, SHEITC (2018-RGZN-02046), and Baidu Scholarship.


\bibliographystyle{IEEEtran}
\bibliography{TIN_PAMI}

\begin{thebibliography}{10}\itemsep=-1pt

\bibitem{visualgenome}
Ranjay Krishna, Yuke Zhu, Oliver Groth, Justin Johnson, Kenji Hata, Joshua Kravitz, Stephanie Chen, Yannis Kalantidis, Li-Jia Li, David~A Shamma, Michael Bernstein, and Li~Fei-Fei.
\newblock Visual genome: Connecting language and vision using crowdsourced dense image annotations.
\newblock In {\em IJCV}, 2016.

\bibitem{Lu2016Visual}
Cewu Lu, Ranjay Krishna, Michael Bernstein, and Fei~Fei Li.
\newblock Visual relationship detection with language priors.
\newblock In {\em ECCV}, 2016.

\bibitem{fang2018weakly}
Hao-Shu Fang, Guansong Lu, Xiaolin Fang, Jianwen Xie, Yu-Wing Tai, and Cewu Lu.
\newblock Weakly and semi-supervised human body part parsing via pose-guided knowledge transfer.
\newblock In {\em CVPR}, 2018.

\bibitem{faster-rcnn}
Shaoqing Ren, Kaiming He, Ross Girshick, and Jian Sun.
\newblock Faster r-cnn: Towards real-time object detection with region proposal networks.
\newblock In {\em NIPS}, 2015.

\bibitem{lu2018beyond}
Cewu Lu, Hao Su, Yonglu Li, Yongyi Lu, Li~Yi, Chi-Keung Tang, and Leonidas~J Guibas.
\newblock Beyond holistic object recognition: Enriching image understanding with part states.
\newblock In {\em CVPR}, 2018.

\bibitem{maskrcnn}
Kaiming He, Georgia Gkioxari, Piotr Doll{\'a}r, and Ross Girshick.
\newblock Mask r-cnn.
\newblock In {\em ICCV}, 2017.

\bibitem{activitynet}
Bernard~Ghanem Fabian Caba~Heilbron, Victor~Escorcia and Juan~Carlos Niebles.
\newblock Activitynet: A large-scale video benchmark for human activity understanding.
\newblock In {\em CVPR}, 2015.

\bibitem{immitation}
Brenna D. Argall, Sonia Chernova, Manuela Veloso, and Brett Browning.
\newblock A survey of robot learning from demonstration.
\newblock In {\em Robotics and autonomous systems}, 2009.

\bibitem{hicodet}
Yu-Wei Chao, Yunfan Liu, Xieyang Liu, Huayi Zeng, and Jia Deng.
\newblock Learning to detect human-object interactions.
\newblock In {\em WACV}, 2018.

\bibitem{Gkioxari2017Detecting}
Georgia Gkioxari, Ross Girshick, Piotr Doll{\'a}r, and Kaiming He.
\newblock Detecting and recognizing human-object interactions.
\newblock In {\em CVPR}, 2018.

\bibitem{qi2018learning}
Siyuan Qi, Wenguan Wang, Baoxiong Jia, Jianbing Shen, and Song-Chun Zhu.
\newblock Learning human-object interactions by graph parsing neural networks.
\newblock In {\em ECCV}, 2018.

\bibitem{gao2018ican}
Chen Gao, Yuliang Zou, and Jia-Bin Huang.
\newblock ican: Instance-centric attention network for human-object interaction detection.
\newblock In {\em arXiv preprint arXiv:1808.10437}, 2018.

\bibitem{vcoco}
Saurabh Gupta and Jitendra Malik.
\newblock Visual semantic role labeling.
\newblock In {\em arXiv preprint arXiv:1505.04474}, 2015.

\bibitem{Maron1998A}
Oded Maron and Tomás Lozano-Pérez.
\newblock A framework for multiple-instance learning.
\newblock In {\em NIPS}, 1998.

\bibitem{hake}
Yong-Lu Li, Liang Xu, Xinpeng Liu, Xijie Huang, Yue Xu, Shiyi Wang, Hao-Shu Fang, Ze Ma, Mingyang Chen, and Cewu Lu.
\newblock PaStaNet: Toward Human Activity Knowledge Engine.
\newblock In {\em CVPR}, 2020.

\bibitem{Sadeghi2012Recognition}
M.~A. Sadeghi and A.~Farhadi.
\newblock Recognition using visual phrases.
\newblock In {\em CVPR}, 2012.

\bibitem{Yatskar2016Situation}
M.~Yatskar, L.~Zettlemoyer, and A.~Farhadi.
\newblock Situation recognition: Visual semantic role labeling for image understanding.
\newblock In {\em CVPR}, 2016.

\bibitem{xu2017scene}
D.~Xu, Y.~Zhu, C.~B.~Choy, and L.~Fei-Fei.
\newblock Scene graph generation by iterative message passing.
\newblock In {\em CVPR}, 2017.

\bibitem{vtranse}
H.~Zhang, Z.~Kyaw, S.-F.~Chang, and T.-S.~Chua.
\newblock Visual translation embedding network for visual relation detection.
\newblock In {\em CVPR}, 2017.

\bibitem{yin2018zoom}
G.~Yin, L.~Sheng, B.~Liu, N.~Yu, X.~Wang, J.~Shao, and C.~C.~Loy.
\newblock Zoom-net: Mining deep feature interactions for visual relationship recognition.
\newblock In {\em arXiv preprint arXiv:1807.04979}, 2018.

\bibitem{yang2018graph}
L.~Yang, L.~Lu, S.~Lee. D.~Batra and D.~Parikh.
\newblock Graph r-cnn for scene graph generation.
\newblock In {\em ECCV}, 2018.

\bibitem{Wang2006Unsupervised}
Y.~Wang, H.~Jiang, Mark.~S.~Drew, Z.-N.~Li, and G.~Mori.
\newblock Unsupervised discovery of action classes.
\newblock In {\em CVPR}, 2006.

\bibitem{Yang2010Recognizing}
W.~Yang, Y.~Wang, and G.~Mori.
\newblock Recognizing human actions from still images with latent poses.
\newblock In {\em CVPR}, 2010.

\bibitem{Ikizler2008Recognizing}
N.~Ikizler, R.~G.~Cinbis, S.~Pehlivan, and P.~Duygulu.
\newblock Recognizing actions from still images.
\newblock In {\em ICPR}, 2008.

\bibitem{hcvrd}
B.~Zhuang, Q.~Wu, C.~Shen, I.~Reid, and A.~v.~d.~Hengel.
\newblock Care about you: towards large-scale human-centric visual relationship detection.
\newblock In {\em arXiv preprint arXiv:1705.09892}, 2017.

\bibitem{Fang2018Pairwise}
H.-S.~Fang, J.~Cao, Y.-W.~Tai, and C.~Lu.
\newblock Pairwise body-part attention for recognizing human-object interactions.
\newblock In {\em ECCV}, 2018.

\bibitem{Delaitre2010Recognizing}
V.~Delaitre, I.~Laptev, and J.~Sivic.
\newblock Recognizing human actions in still images: a study of bag-of-features and part-based representations.
\newblock In {\em BMVC}, 2010.

\bibitem{hico}
Y.~W.~Chao, Z.~Wang, Y.~He, J.~Wang, and J.~Deng.
\newblock Hico: A benchmark for recognizing human-object interactions in images.
\newblock In {\em ICCV}, 2015.

\bibitem{Chao2014Predicting}
C.-Y.~Chen and K.~Grauman.
\newblock Predicting the location of “interactees” in novel human-object interactions.
\newblock In {\em ACCV}, 2014.

\bibitem{Mallya2016Learning}
A.~Mallya and S.~Lazebnik.
\newblock Learning models for actions and person-object interactions with transfer to question answering.
\newblock In {\em ECCV}, 2016.

\bibitem{li2020detailed}
Yong-Lu Li, Xinpeng Liu, Han Lu, Shiyi Wang, Junqi Liu, Jiefeng Li, and Cewu Lu.
\newblock Detailed 2D-3D Joint Representation for Human-Object Interaction.
\newblock In {\em CVPR}, 2020.

\bibitem{kim2020detecting}
D.-J. Kim, X.~Sun, J.~Choi, S.~Lin, and I.~S. Kweon.
\newblock Detecting human-object interactions with action co-occurrence priors.
\newblock In {\em ECCV}, 2020.

\bibitem{hou2020visual}
Z.~Hou, X.~Peng, Y.~Qiao, and D.~Tao.
\newblock Visual compositional learning for human-object interaction detection.
\newblock In {\em ECCV}, 2020.

\bibitem{zhong2020polysemy}
X.~Zhong, C.~Ding, X.~Qu, and D.~Tao.
\newblock Polysemy deciphering network for human-object interaction detection.
\newblock In {\em ECCV}, 2020.

\bibitem{wang2020contextual}
H.~Wang, W.-s. Zheng, and L.~Yingbiao.
\newblock Contextual heterogeneous graph network for human-object interaction detection.
\newblock In {\em ECCV}, 2020.

\bibitem{kim2020uniondet}
B.~Kim, T.~Choi, J.~Kang, and H.~J. Kim.
\newblock Uniondet: Union-level detector towards real-time human-object interaction detection.
\newblock In {\em ECCV}, 2020.

\bibitem{gao2020drg}
C.~Gao, J.~Xu, Y.~Zou, and J.-B. Huang.
\newblock Drg: Dual relation graph for human-object interaction detection.
\newblock In {\em ECCV}, 2020.

\bibitem{liu2020amplifying}
Y.~Liu, Q.~Chen, and A.~Zisserman.
\newblock Amplifying key cues for human-object-interaction detection.
\newblock In {\em ECCV}, 2020.

\bibitem{li2020hoi}
Y.-L. Li, X.~Liu, X.~Wu, Y.~Li, and C.~Lu.
\newblock Hoi analysis: Integrating and decomposing human-object interaction.
\newblock In {\em NeurIPS}, 2020.

\bibitem{Shen2018Scaling}
Liyue Shen, Serena Yeung, Judy Hoffman, Greg Mori, and Li Fei~Fei.
\newblock Scaling human-object interaction recognition through zero-shot learning.
\newblock In {\em WACV}, 2018.

\bibitem{peyre2018detecting}
Julia Peyre, Ivan Laptev, Cordelia Schmid, and Josef Sivic.
\newblock Detecting rare visual relations using analogies.
\newblock In {\em ICCV}, 2019.

\bibitem{Gkioxari2014Actions}
G.~Gkioxari, R.~Girshick, and J.~Malik.
\newblock DActions and attributes from wholes and parts.
\newblock In {\em ICCV}, 2014.

\bibitem{detectron}
R.~Girshick, I.~Radosavovic, G.~Gkioxari, P.~Doll\'{a}r, and K.~He.
\newblock Detectron.
\newblock \url{https://github.com/facebookresearch/detectron}, 2018.

\bibitem{fpn}
T.-Y.~Lin, P.~Doll{\'a}r, R.~B.~Girshick, K.~He, B.~Hariharan, and S.~J.~Belongie.
\newblock Feature pyramid networks for object detection.
\newblock In {\em CVPR}, 2017.

\bibitem{resnet}
K.~He, X.~Zhang, S.~Ren, and J.~Sun.
\newblock Deep residual learning for image recognition.
\newblock In {\em CVPR}, 2016.

\bibitem{fang2017rmpe}
H.-S.~Fang, S.~Xie, Y.-W.~Tai, and C.~Lu.
\newblock {RMPE}: Regional multi-person pose estimation.
\newblock In {\em ICCV}, 2017.

\bibitem{coco}
T.-Y.~Lin, M.~Maire, S.~Belongie, J.~Hays, P.~Perona, D.~Ramanan, P.~Doll{\'a}r, and C.~L.~Zitnick.
\newblock Microsoft coco: Common objects in context.
\newblock In {\em ECCV}, 2014.

\bibitem{devlin2018bert}
Jacob Devlin, Ming-Wei Chang, Kenton Lee, and Kristina Toutanova.
\newblock Bert: Pre-training of deep bidirectional transformers for language understanding.
\newblock In {\em arXiv preprint arXiv:1810.04805}, 2018.

\bibitem{gcn}
T.~N. Kipf and M.~Welling.
\newblock Semi-supervised classification with graph convolutional networks.
\newblock In {\em arXiv preprint arXiv:1609.02907}, 2016.

\end{thebibliography}

%

\begin{IEEEbiography}[{\includegraphics[width=1in,height=1.25in,clip,keepaspectratio]{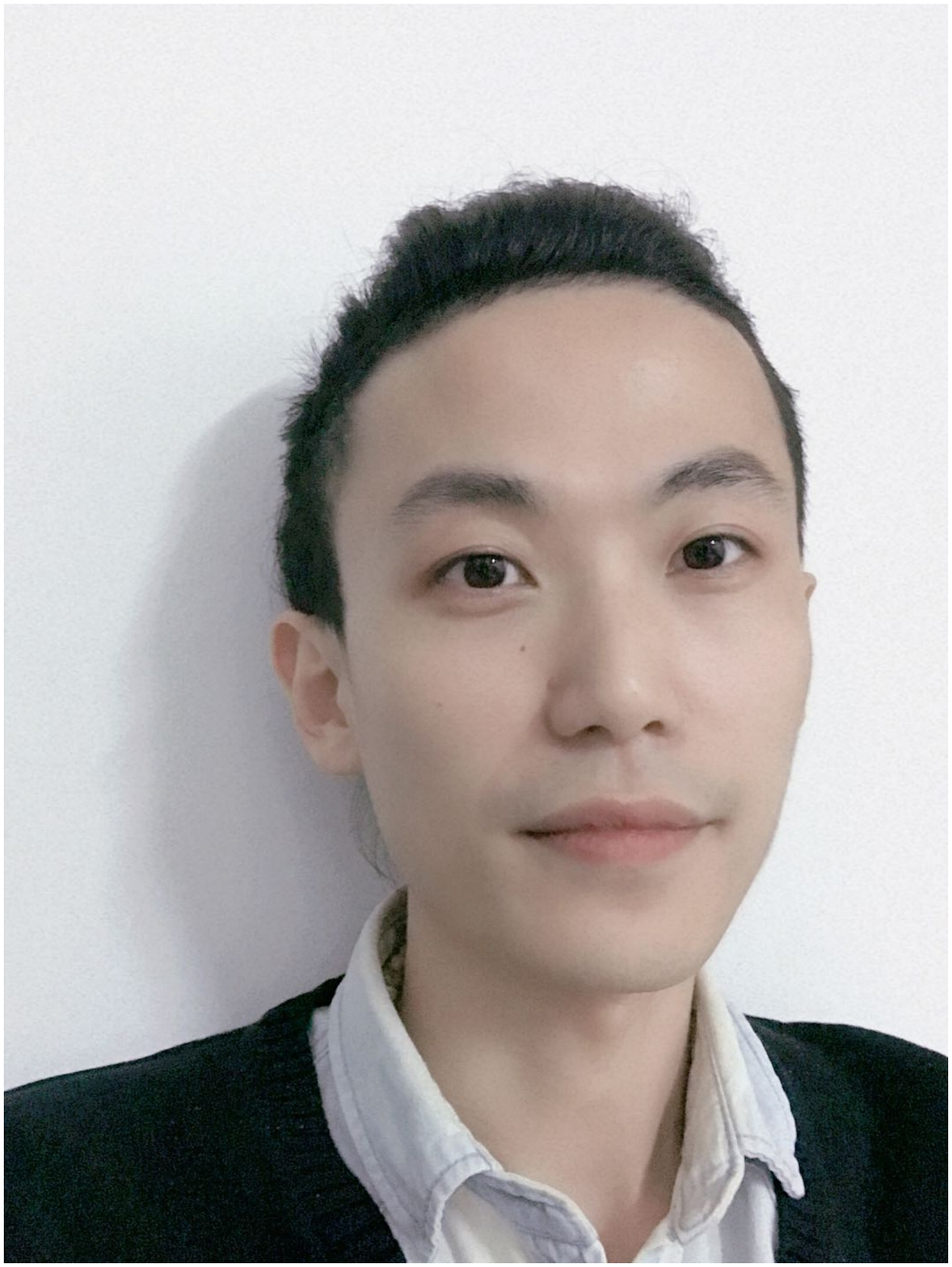}}]{Yong-Lu Li}
received the B.S. degree in automation from Beijing University of Chemical Technology, Beijing, China, in 2012, and the M.S. degree in Control Engineering from Institution of Automation, Chinese Academy of Science and Harbin University of Science and Technology, China, in 2017. He is currently a Ph.D. candidate with MVIG lab, Shanghai Jiao Tong University. His research interests include human activity understanding, knowledge-based visual reasoning, and robotics.
\end{IEEEbiography}

\begin{IEEEbiography}[{\includegraphics[width=1in,height=1.25in,clip,keepaspectratio]{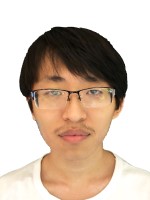}}]{Xinpeng Liu} is currently pursuing the bachelor’s degree at Shanghai Jiao Tong University, Shanghai, China. He is a research assistant at Machine Vision and Intelligence Group, Shanghai Jiao Tong University. His research interests include computer vision and deep learning.
\end{IEEEbiography}


\begin{IEEEbiography}[{\includegraphics[width=1in,height=1.25in,clip,keepaspectratio]{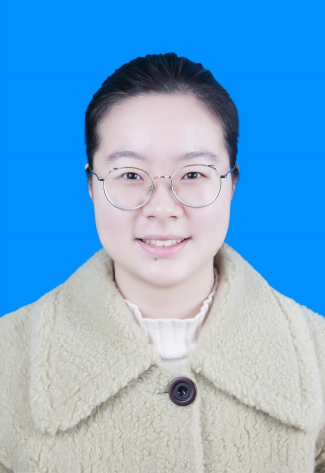}}]{Xiaoqian Wu}
is currently pursuing the bachelor’s degree at Shanghai Jiao Tong University, Shanghai, China. She is a research assistant at Machine Vision and Intelligence Group, Shanghai Jiao Tong University. Her research interests include computer vision and deep learning.
\end{IEEEbiography}

\begin{IEEEbiography}[{\includegraphics[width=1in,height=1.25in,clip,keepaspectratio]{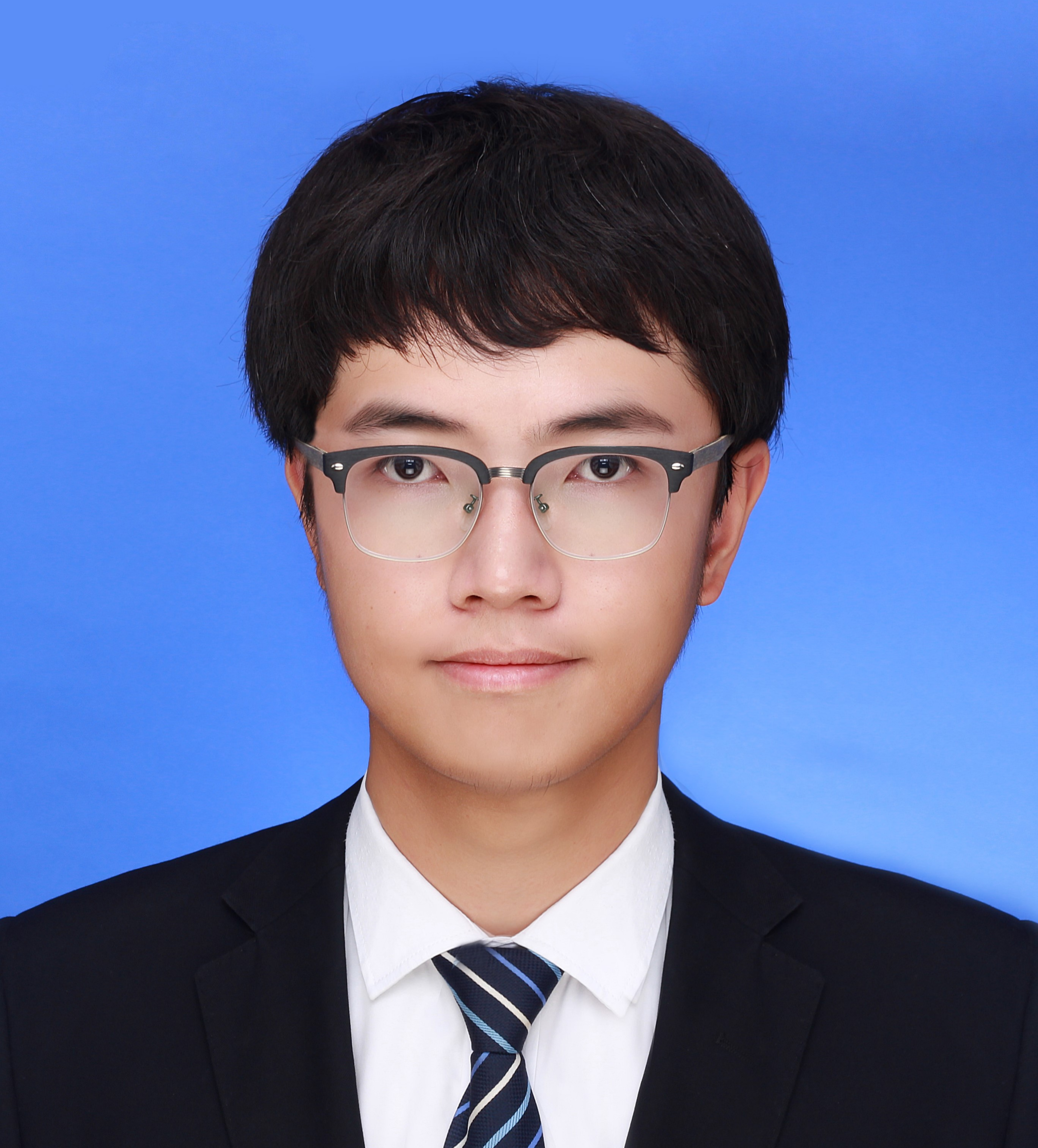}}]{Xijie Huang}
is currently pursuing the bachelor’s degree at Shanghai Jiao Tong University, Shanghai, China. He is a research assistant at Machine Vision and Intelligence Group, Shanghai Jiao Tong University. His research interests include computer vision and deep learning.
\end{IEEEbiography}

\begin{IEEEbiography}[{\includegraphics[width=1in,height=1.25in,clip,keepaspectratio]{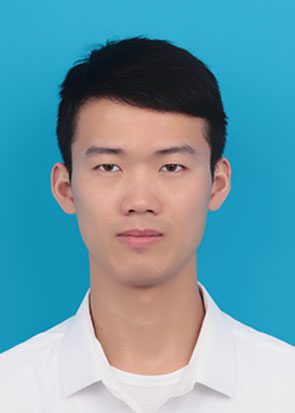}}]{Liang Xu}
received the B.S. degree in computer science at Nanjing University, Nanjing, China in 2018. He is currently a second-year master student at Shanghai Jiao Tong University, supervised by Prof. Cewu Lu. His research interests mainly focus on computer vision.
\end{IEEEbiography}

\begin{IEEEbiography}[{\includegraphics[width=1in,height=1.25in,clip,keepaspectratio]{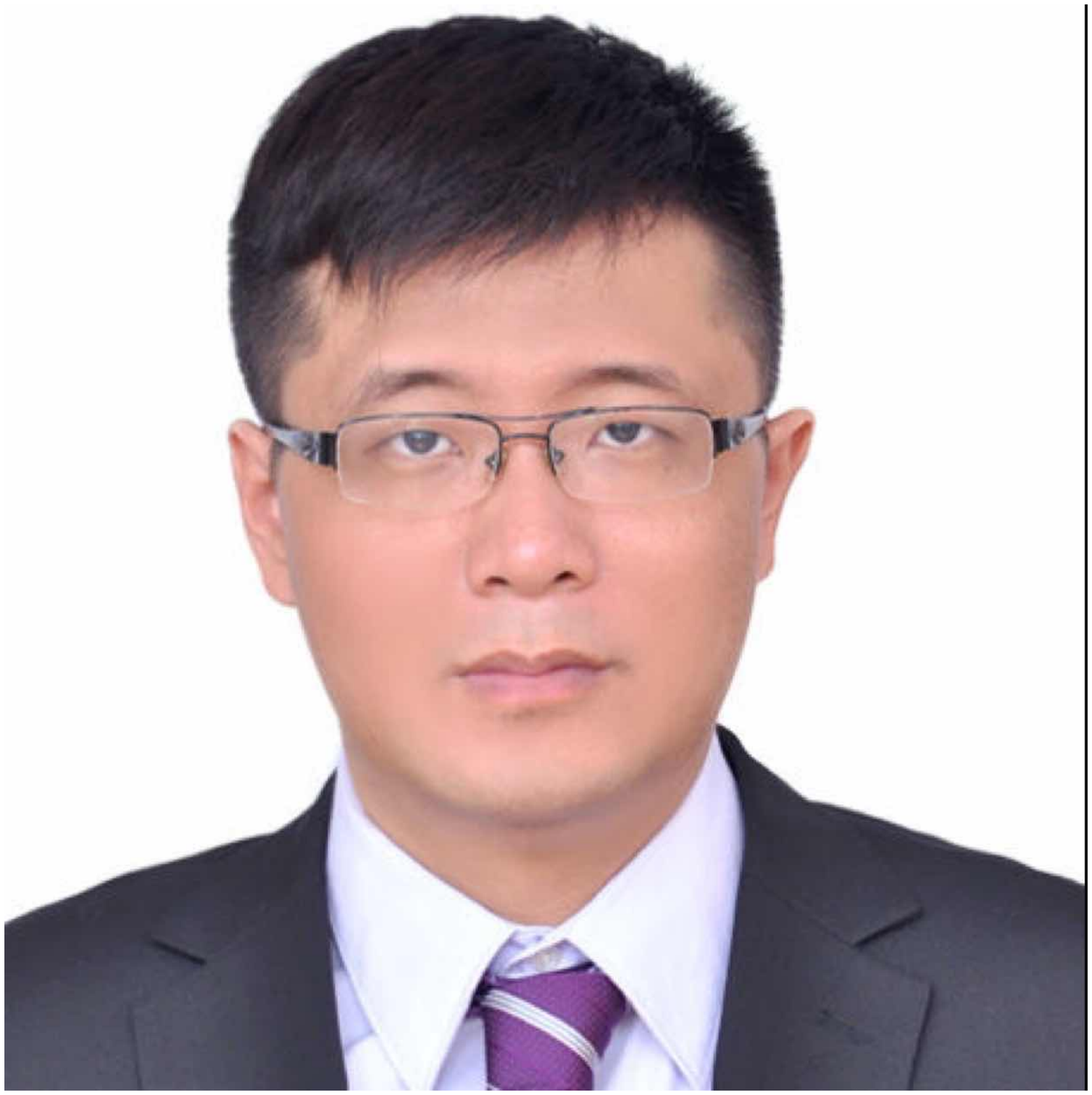}}]{Cewu Lu}
is a Associate Professor at Shanghai Jiao Tong University (SJTU). Before he joined SJTU, he was a research fellow at Stanford University working under Prof. Fei-Fei Li and Prof. Leonidas J. Guibas. He was a Research Assistant Professor at Hong Kong University of Science and Technology with Prof. Chi Keung Tang. He got his Ph.D. degree from The Chinese University of Hong Kong, supervised by Prof. Jiaya Jia. He is one of the core technique members in the Stanford-Toyota autonomous car project. He serves as an associate editor for Journal gtCVPR and reviewer for Journal TPAMI and IJCV. His research interests fall mainly in Computer Vision, deep learning, deep reinforcement learning, and robotics vision.
\end{IEEEbiography}





\newpage




\begin{appendices}
\section{HOI Detection Input Analysis}
\begin{figure}[h]
	\begin{center}
		\includegraphics[width=0.45\textwidth]{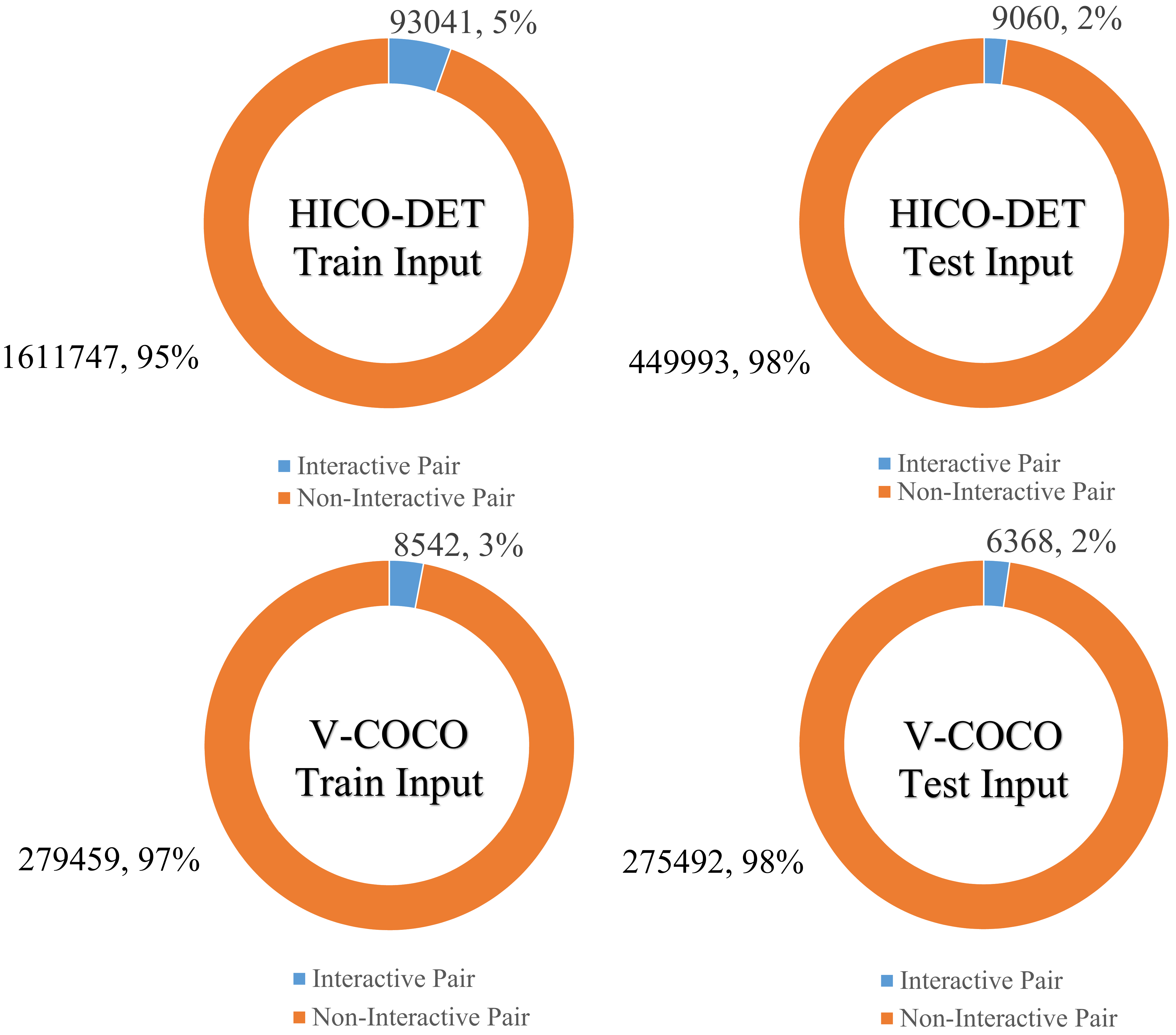}
	\end{center}
	\caption{Input analysis of iCAN~\cite{gao2018ican}.}
	\label{fig:pie}
\end{figure}

In this section, we report the analysis of HOI detection input.
We adopt the object detection results and instance pairing results of iCAN~\cite{gao2018ican}. All pair candidates are produced by an exhaustive pairing of detected humans and objects. We assign pairs as ``interactive'' or ``non-interactive'' according to the HOI ground truth.

As shown in Figure~\ref{fig:pie}, among all pair candidates of HICO-DET~\cite{hicodet} and V-COCO~\cite{vcoco}, non-interactive pairs cover more than 95\% in both training set and test set, which are in the vast majority. 
By leveraging the transferable Interactiveness knowledge learned in the training stage, we can effectively suppress the non-interactive pair candidates in inference using \textbf{Non-Interaction Suppression (NIS)}. Therefore, we do not need to handle such a large number of non-interactive pair candidates.

\section{Pose Feature Encoding}
In this section, we detail pose feature encoding for the pose stream in our network.

\begin{figure}[!ht]
\begin{center}
    \includegraphics[width=0.26\textwidth]{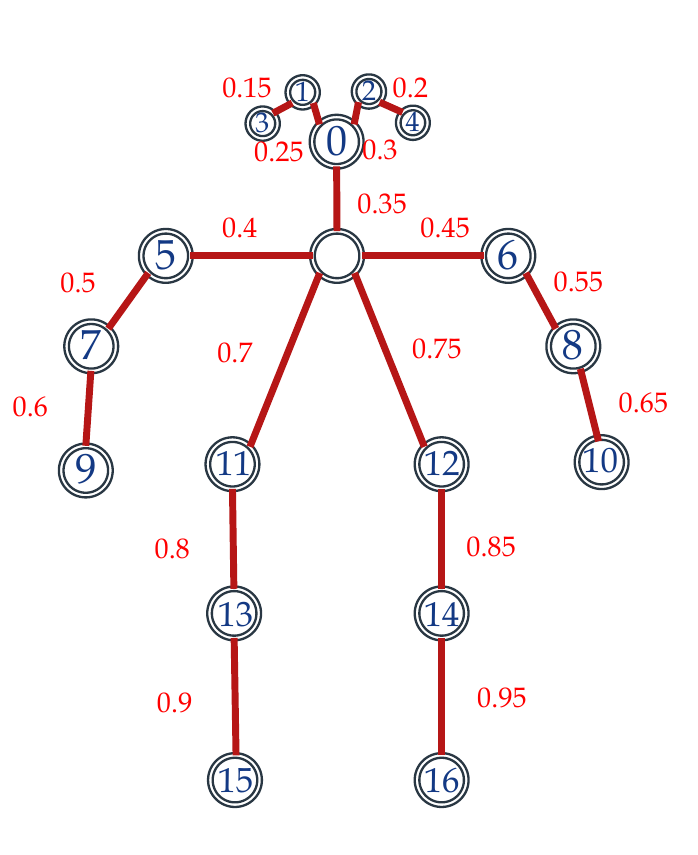}
\end{center}
    \caption{Pose feature encoding. The 17 body keypoints are in COCO format~\cite{coco}, human skeletons connecting keypoints are given different gray values in our setting.}
    \label{fig:pose}
\end{figure}

For the 17 body keypoints in COCO format~\cite{coco}, different body skeletons link two certain keypoints. The skeletons and gray values are shown in Fig.~\ref{fig:pose}.
The pose map is a 64x64 box and each pixel has a gray value between 0 and 1. To distinguish the skeletons, we design their values with an equal interval of 0.05.

\section{PaStaNet-HOI Dataset Analysis}
\begin{figure*}[!htb]
  \centering
  \subfigure[Samples per HOI in the train set of HICO-DET~\cite{hicodet}.]{\includegraphics[width=0.45\textwidth]{{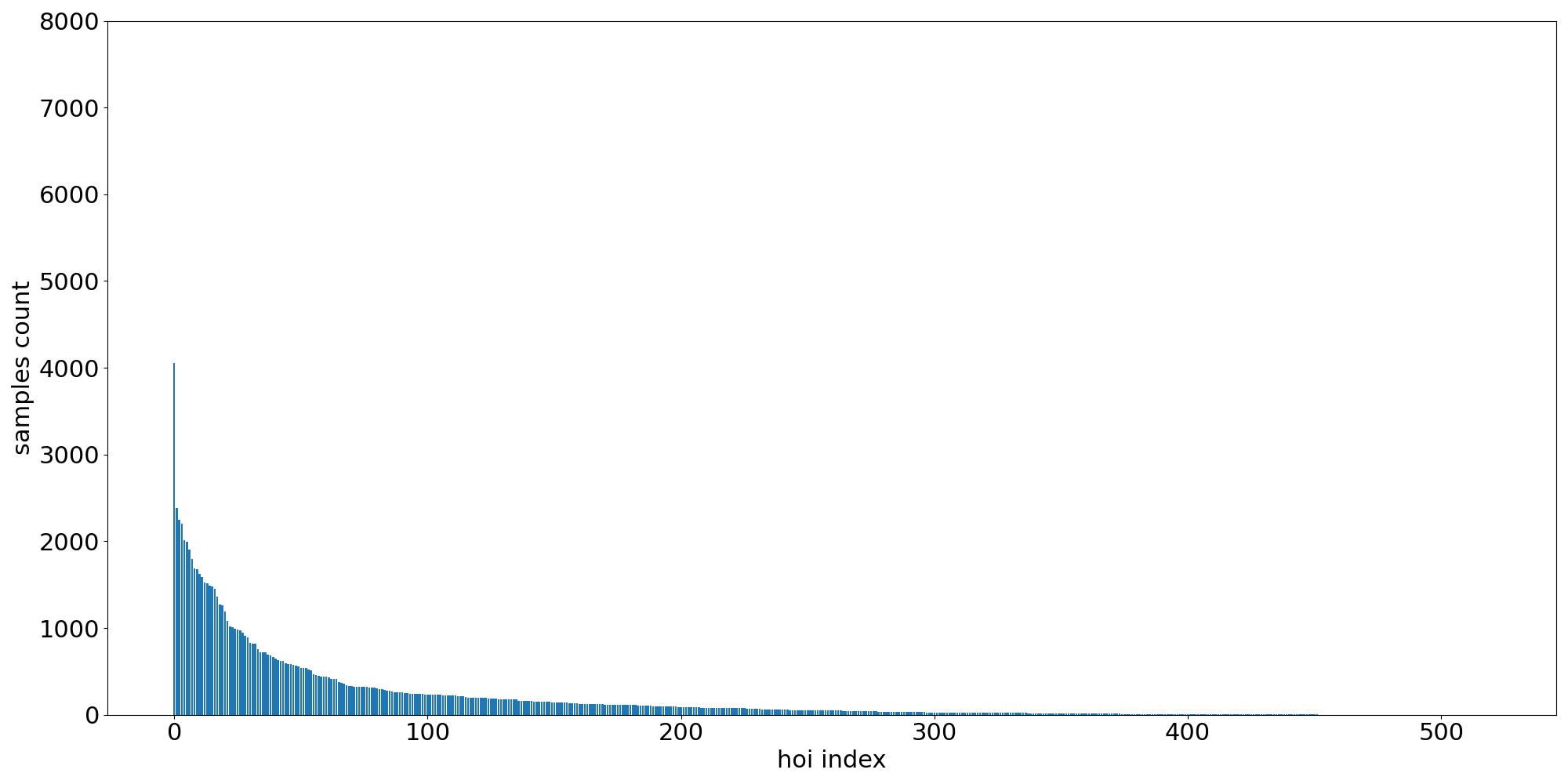}}}
    \subfigure[Samples per HOI in the train set of PaStaNet-HOI~\cite{hake}.]{\includegraphics[width=0.45\textwidth]{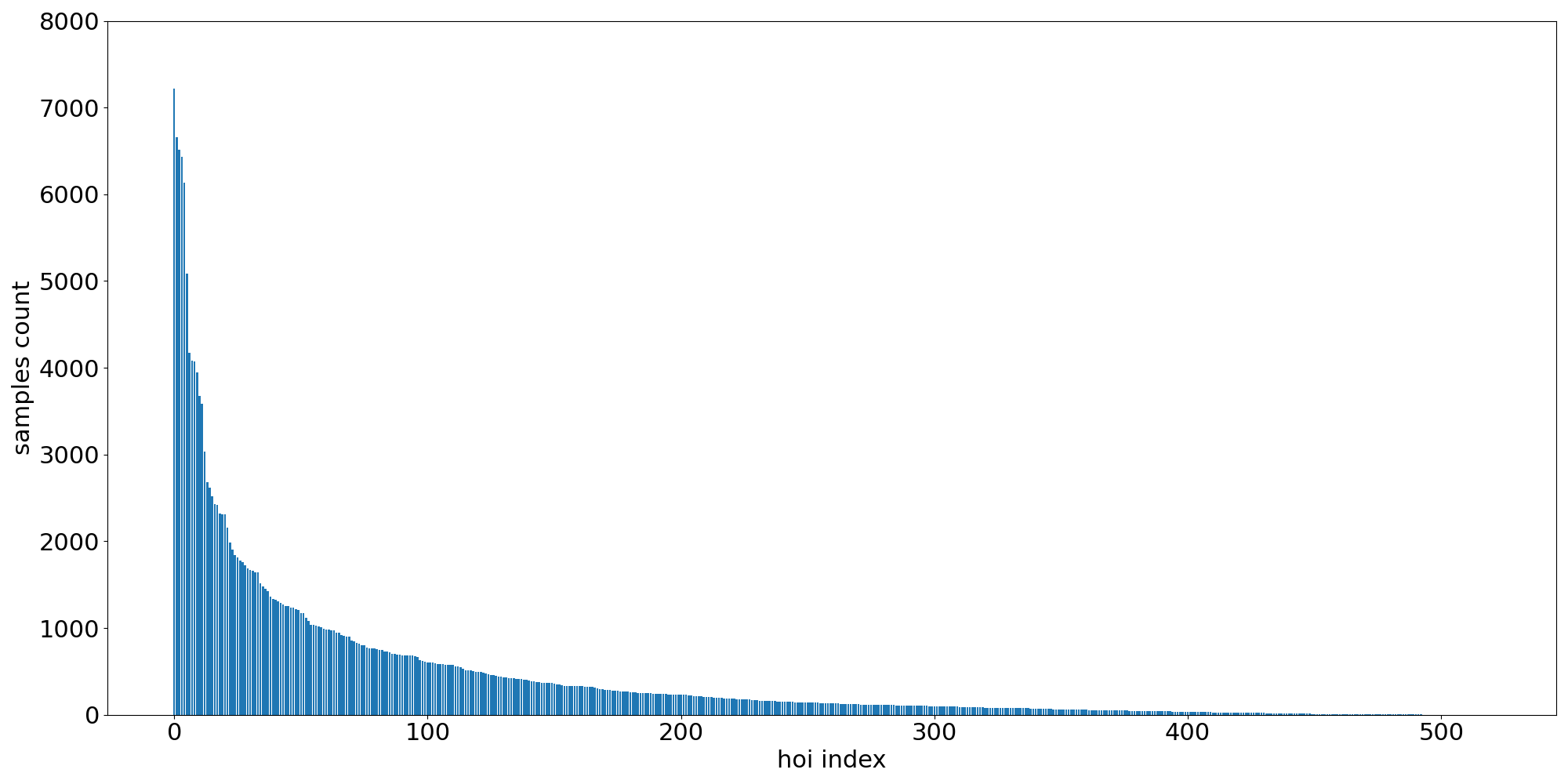}} \\
    \subfigure[Samples per HOI in the test set of HICO-DET~\cite{hicodet}.]{\includegraphics[width=0.45\textwidth]{{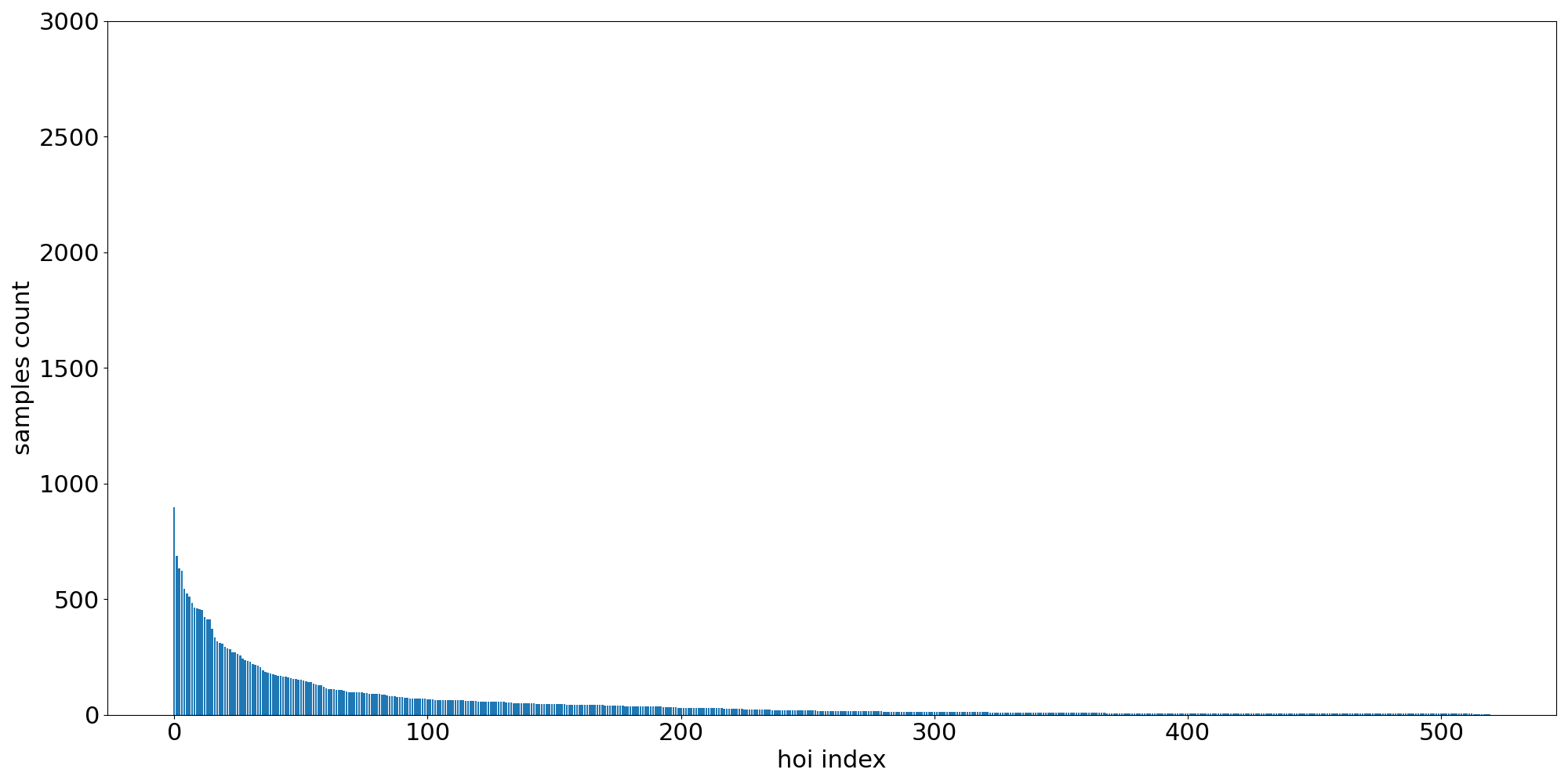}}}
    \subfigure[Samples per HOI in the test set of PaStaNet-HOI~\cite{hake}.]{\includegraphics[width=0.45\textwidth]{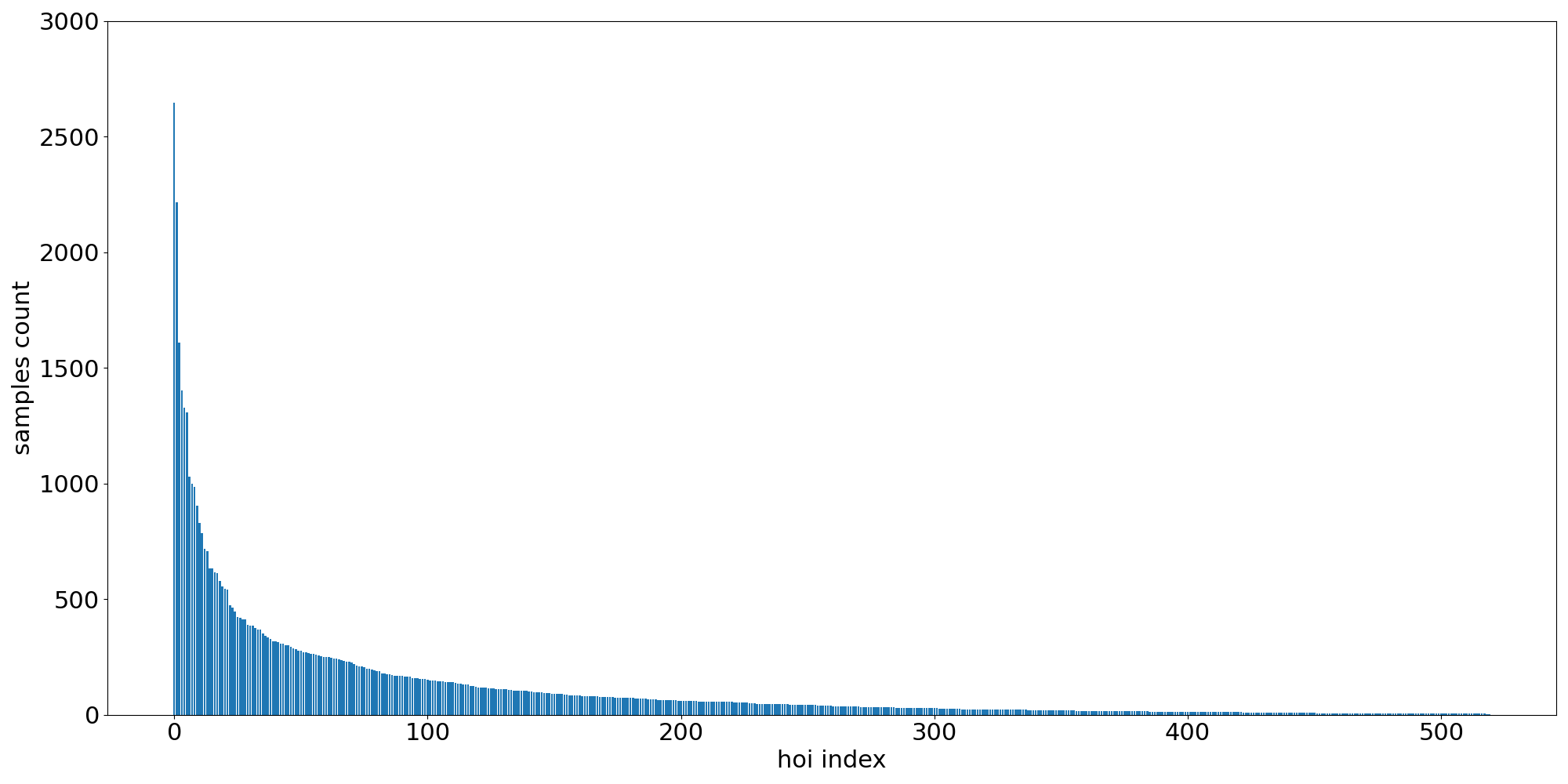}} \\
  \caption{Samples per HOI in the train/test set of HICO-DET~\cite{hicodet}/PaStaNet-HOI.}
    \label{fig:PaStaNet-HOI}
    \vspace{0.2in}
\end{figure*}

We resplit the HOI data of HAKE~\cite{hake} and construct PaStaNet-HOI. 
\textbf{PaStaNet-HOI} provides 110K+ images (77,260 images for train set, 11,298 images for validation set, and 22,156 images for test set). Compared with HICO-DET~\cite{hicodet}, PaStaNet-HOI has larger train and test sets. The interaction categories are similar to the settings of HICO-DET~\cite{hicodet}, but we exclude the 80 ``non-interaction'' categories and only define \textbf{520 HOI categories}. This can help to alleviate the annotation missing in HICO-DET~\cite{hicodet}. 
For these 520 HOI categories, we count the samples in the train/test set of HICO-DET~\cite{hicodet}/PaStaNet-HOI and compare them in Fig.~\ref{fig:PaStaNet-HOI}. 
We can find that PaStaNet-HOI has a more obvious long-tailed distribution. Thus, it is more difficult than HICO-DET, which is also proved by our experiments.

\begin{figure*}[!ht]
	\begin{center}
		\includegraphics[width=0.95\textwidth]{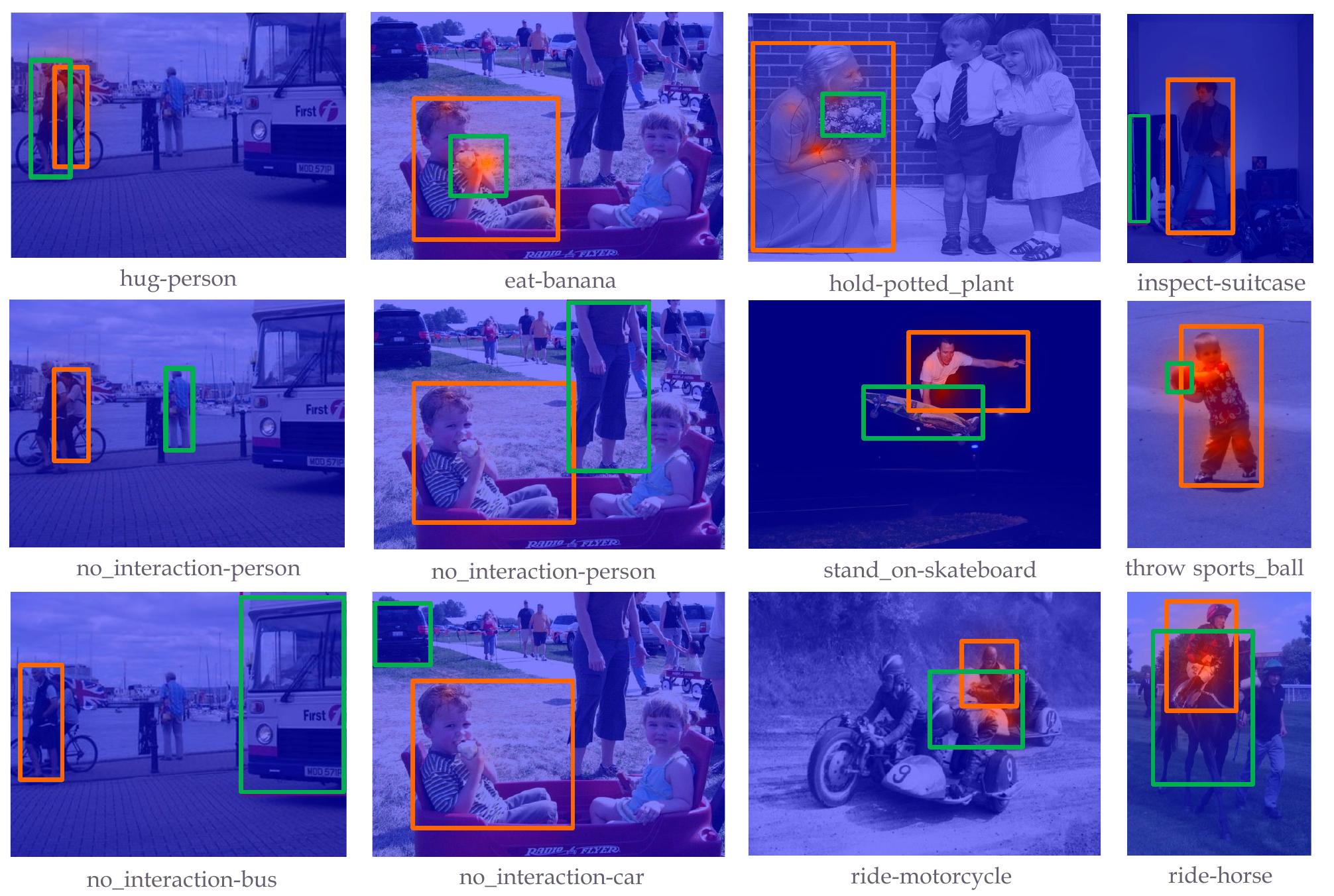}
	\end{center}
	\caption{Additional heatmaps of interactiveness attention based on $\mathbf{R}\mathbf{C}\mathbf{D}_3$ on HICO-DET and $\mathbf{R}\mathbf{C}\mathbf{D}$ on PaStaNet-HOI. The pixels with higher interactiveness probabilities are presented with a brighter color. We can find that part-level interactiveness knowledge localizes the most informative parts effectively.}
	\label{Figure:vis_heatmap_suppl}
\end{figure*}

\section{Additional Heatmaps for Interactiveness Attention} 
We use more visualized samples to further verify the effectiveness of interactiveness attention (Fig.~\ref{Figure:vis_heatmap_suppl}). Especially, more non-HOI samples are given in col1 and col2.

\section{Alternative Relationship Between Part and Whole Body Interactivenesses}
To further explore the relationship between part and whole body interactivenesses, we investigate two alternative methods as follows. 

\subsection{Action-Specific relationship from Language Priors} 
We consider a \textbf{\textit{action-specific}} relationship between part and whole body interactivenesses. In our proposal, we claimed the relationship as: the instance interactiveness is false if and only if all part interactivenesses are false. A human is interacting with an object if and only if at least one body part is interacting with the object. 
However, this is a general relationship instead of an action-specific one. Actually, given an action category, prior knowledge can provide information for \textit{which} body part is more likely to interact with the object. For example, ``ride bicycle'' is more related to feet, thighs, and hands while ``hold apple'' is more related to hands. And these \textbf{closer} relationships can be extracted from language priors.

Specifically, we treat the prior relationship knowledge as an action-part correlation matrix $\mathbf{R_{mn}}=\{r_{ji}\}_{m \times n}$, where $r_{ji} \in [0,1]$ is the correlation between $j$-th action and $i$-th body part. 
$\mathbf{R_{mn}}=\{r_{ji}\}_{m \times n}$ can be generated via the language representations of Bert~\cite{devlin2018bert}. That is, two words would have a larger correlation when their word embeddings are closer. We use the cosine similarity to measure the correlation between two words.
In detail, given the word vector $\mathbf{L}_{pi}$, $\mathbf{L}_{aj}$ of the $i$-th body part and $j$-th action, we can get their correlation as $r_{ji}=\frac{\mathbf{L}_{aj} \cdot \mathbf{L}_{pi}}{\lvert \mathbf{L}_{aj}\rvert \cdot \lvert \mathbf{L}_{pi}\rvert}$. 
With part interactiveness probabilities $\mathbf{Q_{n1}}=\{p_{(p_i,o)}^\mathbf{D}\}_{n \times 1}$, we thus get the \textit{prior} action scores via $\mathcal{S}_{(h,o)}^{'} = \mathbf{S_{m1}}=\mathbf{R_{mn}}\cdot \mathbf{Q_{n1}}$. 
Here, $\mathcal{S}_{(h,o)}^{'}$ indicates the \textbf{prior} action inference relying on part interactiveness predictions and prior action-part correlations only, which would import ``common sense'' into our model. It can be used to modulate the inference of our initial model.

When modulated by $\mathcal{S}_{(h,o)}^{'}$, our interactiveness network (15.38 mAP on PaStaNet-HOI) shows a performance bonus of \textbf{0.87} (multiplication, $\mathcal{S}_{(h,o)}*\mathcal{S}_{(h,o)}^{'}$) and \textbf{0.46} (addition, $\mathcal{S}_{(h,o)}+\mathcal{S}_{(h,o)}^{'}$) mAP on PaStaNet-HOI. It verifies the effectiveness of the action-specific relationship from language priors.
The additional cost is limited since word vectors are beforehand extracted from the pre-trained Bert~\cite{devlin2018bert} base model. Nevertheless, the method is less flexible and sacrifices the transferability, because we need to provide different correlation matrices for different action category settings.

\subsection{Hierarchical Interactiveness Graph} 
\begin{figure}[]
\begin{center}
    \includegraphics[width=0.4\textwidth]{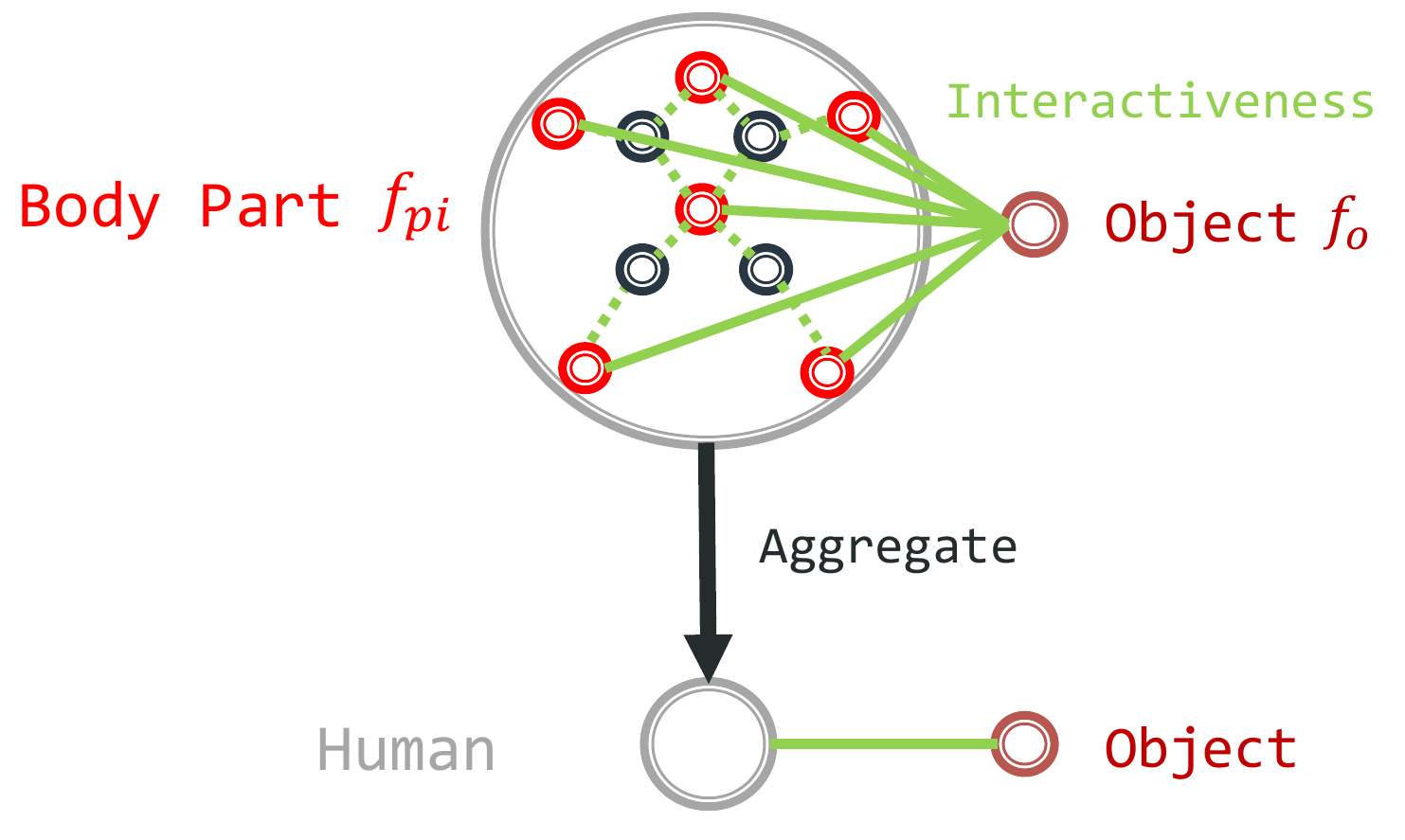}
\end{center}
	\caption{The hierarchical interactiveness graph. To capture the implicit relationships between human body part interactivenesses and whole body interactiveness, we utilize GCN~\cite{gcn} to encode these two levels. The body part interactivenesses are embeded in the edges between body part nodes and object nodes. Meanwhile, after aggregating the 10 part nodes into a human node, we use the edge between human and object nodes as the whole body interactiveness.}
	\label{fig:gcn}
\end{figure}

We also consider an \textbf{\textit{implicit}} relationship between human body part interactivenesses and whole body interactiveness and construct the hierarchical interactiveness graph via GCN (Graph Convolutional Network)~\cite{gcn}. 
As illustrated in Fig.~\ref{fig:gcn}, this graph has two levels.
First, object and ten body parts take $f_o$ and $f_{pi} (1 \le i \le 10)$ as their node features respectively. The body part interactivenesses are embedded in the edges between body part nodes and object nodes. 
Second, after aggregating the 10 part nodes into a human node, we use the edge between human and object nodes as the whole body interactiveness. 
This method gets only 16.15 mAP on Default Full set on HICO-DET~\cite{hicodet}. The possible reason is that implicit hierarchical relationships may be more difficult to learn than the proposed explicit consistency.

\end{appendices}
\end{document}